\documentclass[]{TEAI}
\usepackage{amsmath} 
\usepackage{natbib}
\usepackage{graphicx}
\usepackage{subcaption} 

\usepackage[page,header]{appendix}
\usepackage[utf8]{inputenc} 
\usepackage[T1]{fontenc}    
\usepackage{hyperref}       
\usepackage{url}            
\usepackage{booktabs}       
\usepackage{amsfonts}       
\usepackage{anyfontsize}    
\usepackage{nicefrac}       
\usepackage{microtype}      
\usepackage{wrapfig}

\usepackage{amssymb}  
\usepackage{bbding}  

\usepackage{tikz}  
\usepackage{comment}  
\usepackage{tabularx}  

\usepackage{booktabs}
\usepackage{array}
\usepackage{etoolbox}

\definecolor{lightblue}{RGB}{200, 230, 255}  
\definecolor{headerblue}{RGB}{150, 200, 255} 

\usepackage{pgfplots}
\pgfplotsset{compat=1.18}
\usepackage{xcolor}         
\usepackage{graphicx}
\usepackage{float}
\usepackage{multirow} 
\usepackage{makecell} 
\usepackage{siunitx}  
\usepackage{tikz}
\usepackage{pgf-pie} 
\usepackage{wrapfig}
\usepackage[export]{adjustbox}

\usepackage{ragged2e}      
\usepackage{array}          
\usepackage{caption}        
\usepackage{enumitem}
\usepackage{pifont}
\usepackage[hang,flushmargin]{footmisc} 

\usepackage{tcolorbox}    
\tcbuselibrary{breakable}  
\tcbuselibrary{skins}      

\usepackage{tabularx}
\usepackage{listings}

\newcommand{\queryimg}[1]{\fcolorbox{red}{white}{\includegraphics[width=0.1755\linewidth]{#1}}}
\newcommand{\retimg}[1]{\fcolorbox{blue!70}{white}{\includegraphics[width=0.1755\linewidth]{#1}}}

\title{Attention Itself Could Retrieve.\\ RetrieveVGGT: Training-Free Long Context Streaming 3D Reconstruction via Query-Key Similarity Retrieval}

\makeatletter
\renewcommand{\authorlist}{%
  {\authorfont\bfseries
  Zichen Zou$^{1,2}$, Xiaosong Jia\textsuperscript{\normalfont\Envelope}$^{1,2}$, Zuxuan Wu$^{1,2}$, Yu-Gang Jiang$^{1,2}$}\\[6pt]
  {\affiliationfont
  $^{1}$Institute of Trustworthy Embodied AI (TEAI), Fudan University\\[2pt]
  $^{2}$Shanghai Key Laboratory of Multimodal Embodied AI\\[2pt]
  \textsuperscript{\normalfont\Envelope} Correspondence Author}%
}
\renewcommand{\affiliationlist}{}
\renewcommand{\contributionlist}{}
\makeatother 

\abstract{
Visual Geometry Grounded Transformer (VGGT) advances 3D reconstruction via scalable Transformer architecture, but the quadratic complexity of global attention prevents long context application. StreamVGGT enables streaming with causal attention, yet its KV cache grows linearly with frames, causing memory overflow and quality degradation. 
We present RetrieveVGGT, \textbf{a training-free framework, which formulates context construction for VGGT as a retrieval problem}. By retrieving a fixed number of relevant frames at each step, VGGT maintains a controllable memory budget, which is close to its training context length. Interestingly, we find that the similarity between current frame queries and cached history frame keys at the first global attention layer of VGGT is already a strong indicator of relevance, eliminating the need for additional learned scoring. To enhance information diversity similar to a recommender system, we propose Segment Sampling so that the retrieval spans distinct relevant segments rather than a single high-similarity region. We design a pose-aware spatial memory mechanism that organizes history frames according to their already estimated camera poses, enabling location-aware retrieval. Extensive experiments demonstrate that RetrieveVGGT achieves state-of-the-art performance, outperforming StreamVGGT, TTT3R, and InfiniteVGGT while maintaining constant memory usage regardless of sequence length.  Code is available at \url{https://github.com/zzctmd/RetrieveVGGT}.
}

\begin{document}
\maketitle

\vspace{-1.5em}

\section{Introduction}
\label{sec:intro}

The demand for progressively reconstructing 3D scenes from continuous video streams, i.e., long-context streaming 3D reconstruction, is rapidly increasing in applications such as  autonomous driving~\cite{huang2023tri,zheng2024occworld}, augmented reality~\cite{hong20243d,lei2025mosca,zheng2024gps}, and embodied intelligence~\cite{black2024pi_0,kim2024openvla,li2024cogact,matsuki20254dtam}. Traditional pipelines based on Structure-from-Motion (SfM)~\cite{agarwal2011building,frahm2010building,liu2024robust,schonberger2016structure,wu2013towards,lindenberger2021pixel} and Multi-View Stereo (MVS)~\cite{yao2018mvsnet,yao2019recurrent,furukawa2009accurate,gu2020cascade,chen2019point,habbecke2007surface} are computationally expensive, and sensitive to noise. Recent feed-forward models like VGGT~\cite{wang2025vggt} predict dense geometry end-to-end, achieving impressive generalization but suffering from quadratic complexity of global attention.

Streaming architectures address online reconstruction through various memory mechanisms. Explicit spatial memories (Spann3R~\cite{wang20253d}, Point3R~\cite{wu2025point3r}) anchor past observations in 3D but grow unboundedly. Recurrent state compression (CUT3R~\cite{wang2025continuous}, TTT3R~\cite{chen2025ttt3r}) uses fixed-length states but suffers from catastrophic forgetting. StreamVGGT~\cite{zhuo2025streaming} enables streaming by KV cache under causal attention, yet its cache grows linearly with frames, causing memory overflow and progressive degradation on long sequences. Concurrent to our work, InfiniteVGGT~\cite{yuan2026infinitevggt} prunes the cache to a fixed budget, but its query-agnostic compression discards frame-specific context, limiting reconstruction quality.

\begin{figure}[!t]  
    \centering      
    \includegraphics[width=1.0\linewidth]{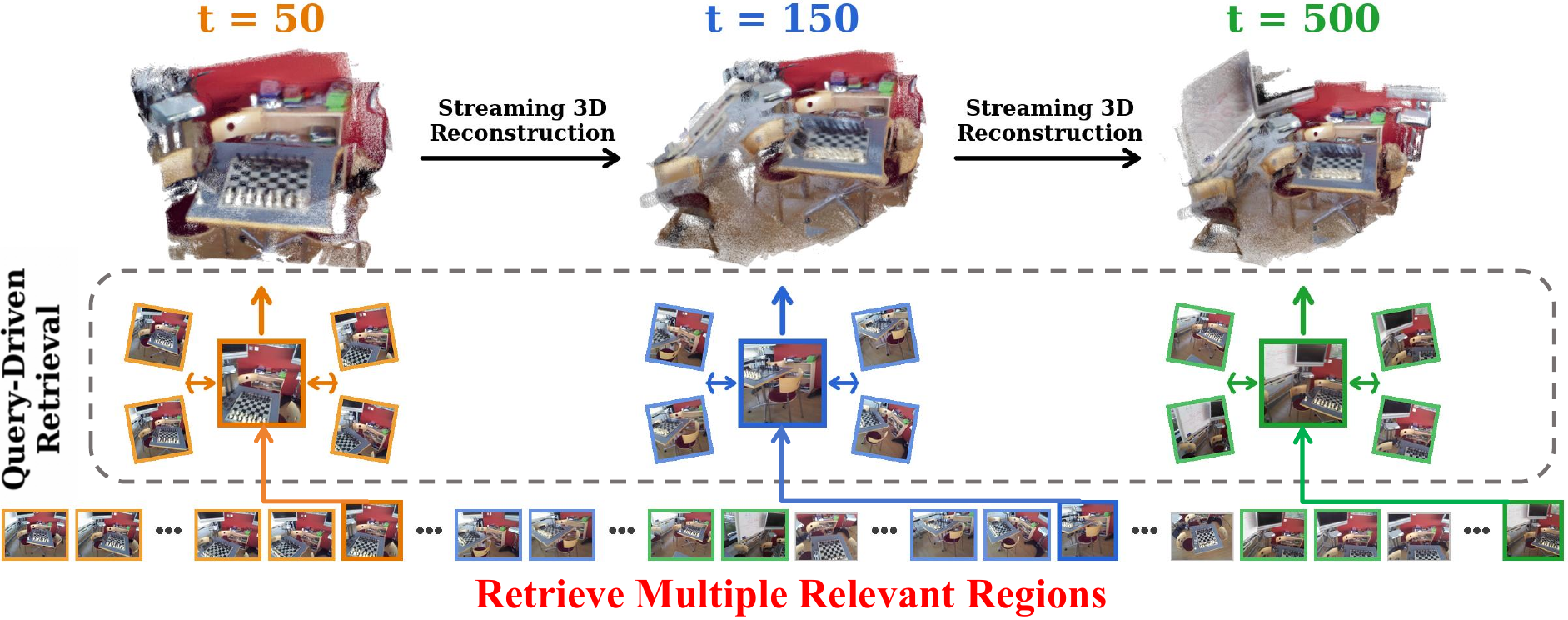}  
    \caption{\textbf{Formulation of RetrieveVGGT}. Given streaming video input (Bottom), RetrieveVGGT retrieves relevant keyframes from history for each incoming query frame via attention-derived relevance (Middle), ensuring flexible and comprehensive scene coverage with long history (Top).
}

    \label{fig:1}
    
\end{figure}

We present \textbf{RetrieveVGGT}, a training-free streaming 3D reconstruction framework, which treats the construction of context for VGGT as a retrieval problem. As in Fig.~\ref{fig:1}, for each incoming frame, RetrieveVGGT retrieves relevant frames from the entire history based on the intrinsic similarity between current frame queries and cached history frame keys at the first global attention layer of VGGT, thereby bounding memory to a fixed budget. To further enhance information diversity much like in a recommender system, we propose Segment Sampling, which identifies and samples from distinct high-relevance segments across the full history. Furthermore, RetrieveVGGT maintains a pose-aware spatial memory that organizes cached KV states by previously estimated camera pose into spatial regions, enabling efficient region-level compression via periodic uniform subsampling for truly scalable long-term memory management.

Our main contributions are summarized as follows:
\begin{itemize}[leftmargin=10pt, topsep=0pt, itemsep=1pt, partopsep=1pt, parsep=1pt,label=$\bullet$]
\item We propose RetrieveVGGT, a retrieval based training-free streaming 3D reconstruction framework that enables each frame to dynamically attend to its most relevant historical keyframes, achieving faithful reconstruction with constant memory cost regardless of sequence length.

\item We introduce Segment Sampling and a pose-aware spatial memory to jointly enhance selection diversity across the full history and scalable long-term memory management.

\item Extensive experiments on 3D reconstruction, video depth estimation, and camera pose estimation demonstrate that RetrieveVGGT achieves state-of-the-art performance with up to 20\% improvement over existing streaming methods, while maintaining bounded GPU memory consumption.
\end{itemize}

\section{Related Work}

\subsection{Classic 3D Reconstruction.}
\vspace{-1mm}
  Recovering 3D structure from multi-view images has traditionally relied on geometric optimization and rendering techniques. Structure-from-Motion (SfM)~\cite{agarwal2011building,frahm2010building,liu2024robust,schonberger2016structure,wu2013towards,lindenberger2021pixel} reconstructs sparse geometry and camera poses through feature extraction~\cite{dusmanu2019d2,lowe2004distinctive,rublee2011orb}, matching~\cite{chen2021learning,lindenberger2023lightglue,shi2022clustergnn,wu2013towards}, triangulation, and bundle adjustment~\cite{agarwal2010bundle,triggs1999bundle}, with COLMAP~\cite{schonberger2016structure} a representative system, yet both SfM and Multi-View Stereo (MVS)~\cite{yao2018mvsnet,yao2019recurrent,furukawa2009accurate,gu2020cascade,chen2019point,habbecke2007surface} remain multi-stage, tightly coupled, and offline. Neural Radiance Fields (NeRF)~\cite{barron2022mip,chen2022tensorf,mildenhall2021nerf,zhang2020nerf++,wang2021neus} and 3D Gaussian Splatting (3DGS)~\cite{kerbl20233d,charatan2024pixelsplat,xiang2025gaussianroom,yuan2025robust} achieve photorealistic novel view synthesis, yet demand scene-specific training and cannot generalize to unseen environments at test time. Simultaneous Localization and Mapping (SLAM)~\cite{davison2007monoslam,engel2014lsd,klein2007parallel,newcombe2011dtam} offers online joint localization and mapping, but often depends on specific motion assumptions or auxiliary sensors~\cite{newcombe2011kinectfusion} and produces incomplete sparse maps. These shortcomings motivate end-to-end learning-based approaches.

\subsection{Feedforward 3D Reconstruction.}
\vspace{-1mm}
A paradigm shift began with DUSt3R~\cite{wang2024dust3r}, which formulates reconstruction as pointmap regression from image pairs via a Vision Transformer, unifying matching, pose estimation, and geometry prediction end-to-end. MASt3R~\cite{leroy2024grounding} improves metric accuracy with pixel-level correspondences, and MonST3R~\cite{zhang2024monst3r} adapts the formulation to dynamic scenes, yet these pairwise methods still require costly global alignment when handling more than two views. To eliminate this bottleneck, multi-view feed-forward architectures have emerged: Fast3R~\cite{yang2025fast3r} extends pairwise regression to N-view inputs with parallel view fusion, while VGGT~\cite{wang2025vggt} scales to a 1.2B-parameter transformer that jointly predicts poses, depth, pointmaps, achieving state-of-the-art accuracy across diverse 3D tasks. FastVGGT~\cite{shen2025fastvggt} reduces its cost through training-free token merging, and VGGTLong~\cite{deng2025vggt} decomposes long trajectories into sub-maps for extended capacity. The quadratic complexity of global attention limits these methods to short, offline settings, motivating streaming architectures.

\subsection{Streaming Online 3D Reconstruction.}
\vspace{-1mm}
Streaming architectures process frames incrementally. Explicit spatial memory methods (Spann3R~\cite{wang20253d}, Point3R~\cite{wu2025point3r}) anchor past observations at 3D positions, yet their memory grows unboundedly with scene extent and requires heuristic pruning once saturated. Recurrent state methods (CUT3R~\cite{wang2025continuous}, TTT3R~\cite{chen2025ttt3r}, TTSA3R~\cite{zheng2026ttsa3r}, MUT3R~\cite{shen2025mut3r}) compress history into a fixed-length representation, yet suffer from information loss over long sequences or remain bounded by limited state capacity. Causal attention methods (STream3R~\cite{lan2025stream3r}, StreamVGGT~\cite{zhuo2025streaming}) cache key-value states under causal attention, yet the KV cache grows unboundedly on long inputs, leading to memory overflow and redundancy. Concurrent to our work, InfiniteVGGT~\cite{yuan2026infinitevggt} proposes training-free token pruning to maintain a fixed-budget cache, though such pruning is query-agnostic regardless of per-frame needs. In contrast, RetrieveVGGT adopts a retrieval-based paradigm where the context each frame attends to is explicitly and dynamically adjusted based on its relevance profile.

\begin{figure}[tb]  
    \centering      
    \includegraphics[width=1\linewidth]{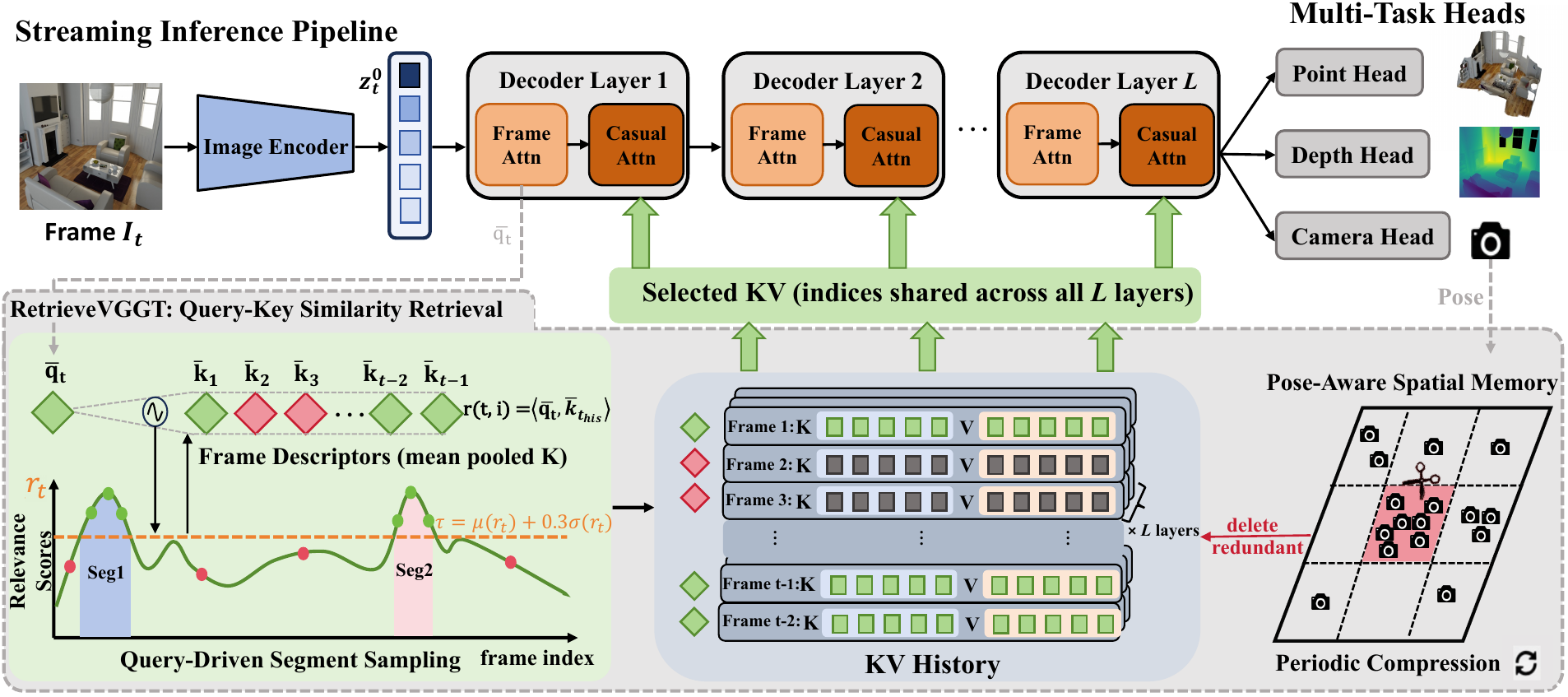}  
    \caption{\textbf{Retrieval Mechanism of RetrieveVGGT.}
For each frame $I_t$:
\textbf{(1) Query-Driven Frame Selection}: mean pooled query is compared against all history frame descriptors at the first attention layer, requiring no extra learned module.
\textbf{(2) Segment Sampling}: selects relevant frames across distinct segments to avoid over-concentration. KV history at the selected indices is reused across all $L$ layers.
\textbf{(3) Pose-Aware Spatial Memory}: organizes KV states by estimated camera poses and periodically compresses over-populated regions to bound memory with spatial diversity.
}
    \label{fig:2}

\end{figure}
\section{Method}
\vspace{-1mm}

RetrieveVGGT is a training-free streaming 3D reconstruction framework built upon StreamVGGT~\cite{zhuo2025streaming} that dynamically selects relevant historical keyframes for each incoming query frame, as in Fig.~\ref{fig:2}. We begin by introducing the StreamVGGT inference pipeline as our foundation (Sec.~\ref{sec:preliminary}), then present the query-driven frame selection mechanism (Sec.~\ref{sec:query_driven}), followed by two complementary components: Segment Sampling for diverse keyframe coverage (Sec.~\ref{sec:segment_sampling}) and pose-aware spatial memory for long-term compression (Sec.~\ref{sec:spatial_compression}).

\subsection{Preliminary: StreamVGGT}
\label{sec:preliminary}
\begin{figure}[tb]  
    \centering      
    \includegraphics[width=1\linewidth]{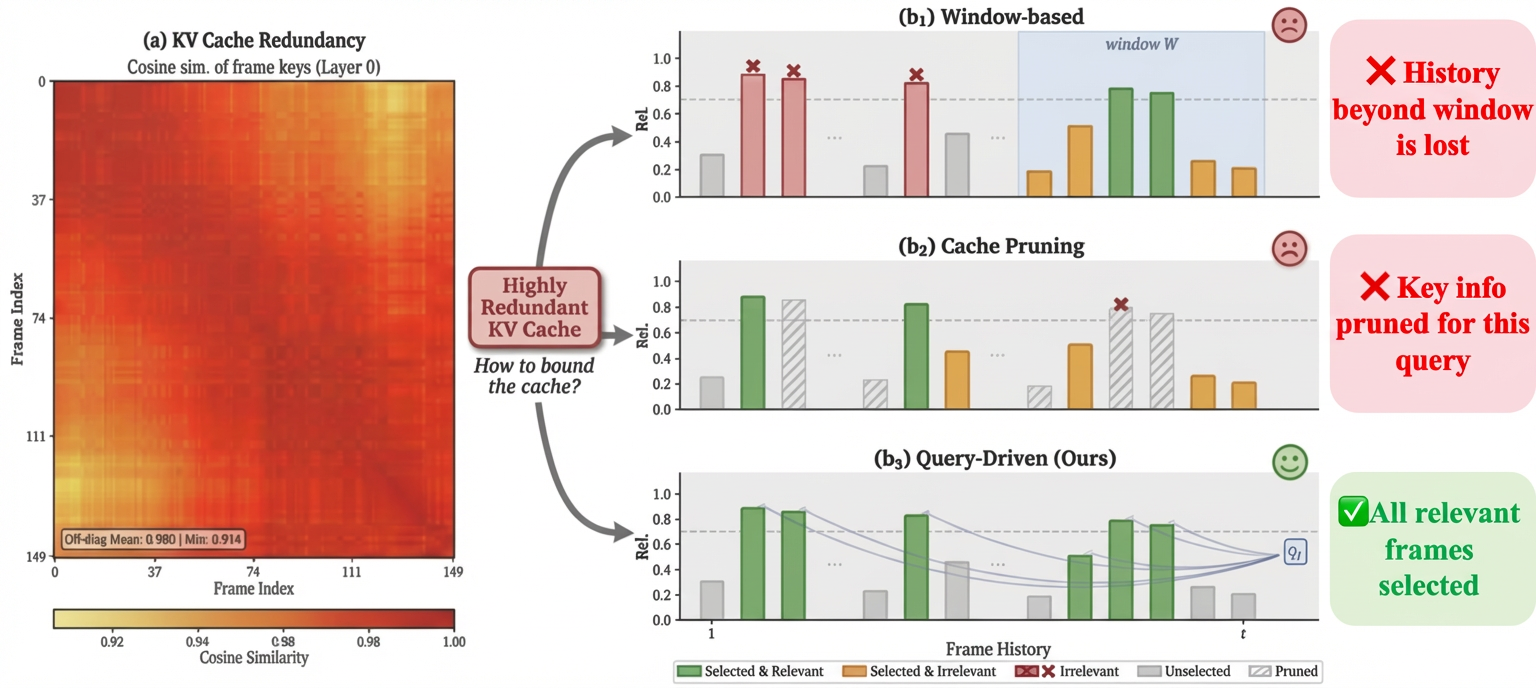}  
    \caption{\textbf{Analysis of Cache and Retrieval Strategy of History Frames.}
        \textbf{(a)}~Cosine similarity of key descriptors across 150 frames reveals \textbf{high redundancy in KV cache} (mean similarity $\approx 0.98$), motivating subset selection.
        \textbf{(b)}~Comparison of three cache management approaches:
        \textbf{(b\textsubscript{1})}~Window-based methods lose all information outside their window;
        \textbf{(b\textsubscript{2})}~Cache pruning maintains a globally compressed cache not tailored to each query, losing relevant information;
        \textbf{(b\textsubscript{3})}~RetrieveVGGT (Ours) dynamically selects the most relevant frames from the entire history based on the current query~$q_t$.
    }
    \label{fig:3}

\end{figure}

\vspace{-1mm}
StreamVGGT is based on VGGT~\cite{wang2025vggt}, consisting of an image encoder $\mathcal{E}$, $L$ transformer aggregator layers, and prediction heads. Given a new frame $I_t$ at time step $t$, the image encoder extracts a token sequence $\mathbf{z}_t^{0}  = \mathcal{E}(I_t)$. The tokens then pass through the $L$ transformer aggregator layers. At each layer $l$, a frame attention block first refines the tokens per image $\hat{\mathbf{z}}_t^{l} = \text{FrameAttn}^{l}(\mathbf{z}_t^{l-1})$. Then, it passes through causal attention block $\mathbf{z}_t^{l} = \text{CausalAttn}^{l}\!\left(\hat{\mathbf{z}}_t^{l},\; \mathbf{K}_{1:t}^{l},\; \mathbf{V}_{1:t}^{l}\right)$ where $\mathbf{K}_{1:t}^{l}$ and $\mathbf{V}_{1:t}^{l}$ denote the KV cache from history frames $1$ through $t$ at layer $l$. Finally, the prediction heads utilize certain layer features for outputs. We could observe that the KV cache grows linearly with the sequence length $t$ and this unbounded growth leads to GPU memory overflow and introduces substantial information redundancy that degrades reconstruction quality, as in Fig.~\ref{fig:3}(a). RetrieveVGGT addresses this by replacing the full KV cache with a retrieval mechanism for relevant frames, described in next section.

\vspace{-1mm}
\subsection{Query-Driven Frame Selection}
\vspace{-1mm}
\label{sec:query_driven}
A solution to the unbounded KV growth is to limit which historical states each frame attends to. Window-based methods (e.g., WinT3R~\cite{li2025wint3r}) restrict attention to a local temporal neighborhood, but lose  earlier context (Fig.~\ref{fig:3}($b_1$)). Cache pruning methods (e.g., concurrent work InfiniteVGGT~\cite{yuan2026infinitevggt}) maintain a globally compressed cache within a fixed budget, yet all future frames share the same pruned context regardless of viewpoint, losing relevant information (Fig.~\ref{fig:3}($b_2$)). RetrieveVGGT instead adopts a \textit{query-driven} paradigm (Fig.~\ref{fig:3}($b_3$)): each incoming frame dynamically selects  relevant keyframes from the \textit{entire} history based on its own query, bounding memory to a fixed budget while preserving long-range coverage tailored to each frame's actual need.
Concretely, at time step $t$, frame $I_t$ passes through the image encoder and frame attention at layer 0, yielding $\hat{\mathbf{z}}_t^{0} \in \mathbb{R}^{P \times d}$ where $P$ is the number of patches per image. The \textbf{first} global attention block of VGGT (no interaction with history before this layer) computes Query, Key, Value for current frame:

\begin{equation}
\mathbf{Q}_t^{0},\; \mathbf{K}_t^{0},\; \mathbf{V}_t^{0} = \text{Proj}_{qkv}^{0}\!\left(\text{LN}^{0}(\hat{\mathbf{z}}_t^{0})\right) \in \mathbb{R}^{H \times P \times d_h},
\end{equation}
where $H$ is the number of heads and $d_h = d / H$. Each token sequence begins with $s$ special tokens (camera and register tokens) followed by $P_p = P - s$ patch tokens. Since only patch tokens encode frame-specific visual content, we mean-pool over patch positions for frame-level descriptors:
\begin{equation}
\bar{\mathbf{q}}_t = \frac{1}{P_p}\sum_{j=s+1}^{P} \mathbf{Q}_{t}^{0}[:,j,:] \in \mathbb{R}^{H \times d_h}, \quad
\bar{\mathbf{k}}_t = \frac{1}{P_p}\sum_{j=s+1}^{P} \mathbf{K}_{t}^{0}[:,j,:] \in \mathbb{R}^{H \times d_h},
\end{equation}
where $\bar{\mathbf{q}}_t$ and $\bar{\mathbf{k}}_t$ are frame-level query and key descriptors of the current frame, respectively; only patch tokens (indices $s{+}1$ to $P$) are averaged, excluding $s$ special tokens. The key descriptor $\bar{\mathbf{k}}_t$ is cached for retrieval. We compute the relevance score between the current query $\bar{\mathbf{q}}_t$ and each history descriptor $\bar{\mathbf{k}}_i$:
\vspace{-3mm}
\begin{equation}
r(t, i) =  \frac{1}{H}\sum_{h=1}^{H} \langle \bar{\mathbf{q}}_t[h,:],\; \bar{\mathbf{k}}_{i}[h,:] \rangle, \quad i = 1, \dots, t{-}1.
\end{equation}

For history frame selection, given a frame budget $N$, we always reserve the first frame since it is defined as the coordinate reference in VGGT~\cite{wang2025vggt}.  The remaining $N{-}1$ frames are selected via Segment Sampling (Sec.~\ref{sec:segment_sampling}) based on the relevance scores $\mathbf{r}_t$, producing the selected set $\mathcal{S}_t$. Selection operates at the frame level to preserve within-frame structural integrity, and is performed only at layer 0 and then reused across all $L$ layers, as we empirically observe no enhancement for layer-wise selection but with  extra latency. At each layer $l$, the retrieved frames' KV are used:
\begin{equation}
\mathbf{z}_t^{l} = \text{Attn}\!\left(\mathbf{Q}_t^{l},\; \mathbf{K}_{\mathcal{S}_t}^{l},\; \mathbf{V}_{\mathcal{S}_t}^{l}\right).
\end{equation}
After all layers, aggregated features pass through task-specific heads to produce 3D predictions. The estimated camera pose of current frame and descriptor $\bar{\mathbf{k}}_t$ are recorded in spatial memory (Sec.~\ref{sec:spatial_compression}).

\subsection{Segment Sampling}

\label{sec:segment_sampling}

\begin{figure}[tb]  
    \centering      
    \includegraphics[width=1.0\linewidth]{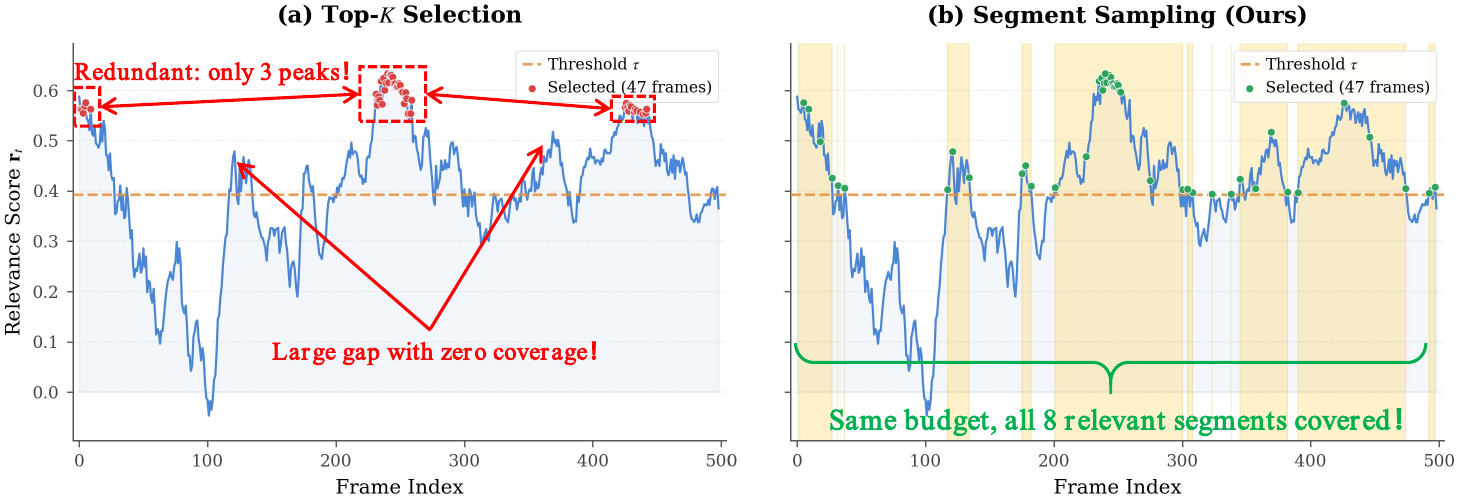}  
    \caption{%
    \textbf{Comparison of Frame Selection}. on the \texttt{office/seq-07} sequence of 7-Scenes (500 frames, $K{=}47$).
    \textbf{(a)}~Top-$K$ clusters frames around a single peak, covering only 3/8 high-relevance segments.
    \textbf{(b)}~Segment Sampling (Ours) distributes frames across 8 segments above threshold~$\tau$ (yellow regions), reducing Accuracy error from 0.281\,m to 0.036\,m (\textbf{7.8$\times$}).
    }
    \label{fig:segment_sampling}
    \vspace{-4mm}

\end{figure}

As in Fig.~\ref{fig:segment_sampling} (a), a naïve top-$K$ selection on the relevance scores $\mathbf{r}_t$ tends to concentrate all selected frames within a single high-similarity peak, producing redundant views of one scene region while neglecting other informative parts of the history. Similar to the \textbf{retrieval diversity} in a recommender system,  as in  Fig.~\ref{fig:segment_sampling} (b), Segment Sampling addresses this by identifying \textit{multiple} high-relevance segments and distributing the frame budget proportionally among them. Qualitative examples of retrieved keyframes, confirming that Q-K similarity indeed selects geometrically relevant views, are in Appendix A. The pipeline of Segment Sampling decomposes frame selection into four sequential steps, each designed to address a specific challenge arising from the previous stage.

\noindent\textbf{Segment Identification.}
We first compute an adaptive threshold based on relevance scores across all history frames $\mathbf{r}_t$:
\begin{equation}
  \tau = \mu(\mathbf{r}_t) + w_{\text{thre}} \,\sigma(\mathbf{r}_t),
\end{equation}
where $\mu$ and $\sigma$ denote mean and standard deviation. $ w_{\text{thre}}$ (set to 0.3) is a hyperparameter, where a frame is relevant if its score exceeds the mean by more than $w_{\text{thre}}$ times the standard deviation. 

Then, a \textit{segment} is defined as a maximal contiguous subsequence of frames whose relevance scores all exceed $\tau$. Adjacent segments separated by fewer than $\delta$ frames are automatically merged together to avoid excessive over-fragmentation, yielding $M$ distinct coherent segments $\{S_1, S_2, \dots, S_M\}$.

\noindent\textbf{Proportional Quota Allocation.}
Let $r_k^{\max} = \max_{i \in S_k} r(t, i)$ be the peak relevance within segment $S_k$. The frame budget $N_{\text{sel}}$ is first distributed in proportion to peak importance:
\begin{equation}
  \hat{n}_k = \left\lfloor N_{\text{sel}} \cdot \frac{r_k^{\max}}{\sum_{j=1}^{M} r_j^{\max}} \right\rfloor.
\end{equation}
The raw quota $\hat{n}_k$ is then clamped to ensure feasibility:
\begin{equation}
  n_k = \mathrm{clamp}\!\left(\hat{n}_k,\;1,\;|S_k| \right),
\end{equation}
where the lower bound $1$ guarantees that every detected segment contributes at least one frame, while $|S_k|$ prevents over-sampling beyond its own segment length.

\noindent\textbf{Within-Segment Sampling.}
Within each segment $S_k$, we first select the peak frame $i_k^{*} = \arg\max_{i \in S_k} r(t, i)$, then uniformly sample the remaining $n_k - 1$ frames across the segment span:
\begin{equation}
  \mathcal{F}_k = \bigl\{i_k^{*}\bigr\} \;\cup\; \operatorname{Uniform}\!\bigl(S_k \setminus \{i_k^{*}\},\; n_k - 1\bigr),
\end{equation}
where $\operatorname{Uniform}(S, n)$ draws $n$ evenly spaced frames from set $S$.

\noindent\textbf{Budget Adjustment.}
After collecting all samples $\mathcal{F}_{\text{seg}} = \bigcup_{k=1}^{M} \mathcal{F}_k$ from all segments, the total count may deviate significantly from $N_{\text{sel}}$. When the total number of detected segments $M$ exceeds $N_{\text{sel}}$, the guaranteed minimum of one frame per segment necessarily pushes $|\mathcal{F}_{\text{seg}}|$ above $N_{\text{sel}}$; conversely, rounding and the strict upper bounds ($|S_k|$) may truncate quotas substantially below their proportional share, thereby leaving $|\mathcal{F}_{\text{seg}}| < N_{\text{sel}}$. We reconcile each case as follows:

When the budget is exceeded, we retain only the $N_{\text{sel}}$ highest-scoring samples:
\begin{equation}
  \mathcal{F}_{\text{final}} = \operatorname{Top}\!\bigl(N_{\text{sel}},\;\mathcal{F}_{\text{seg}}\bigr).
\end{equation}

When a surplus remains, we fill the gap with highest-scoring unselected frames:
\begin{equation}
  \mathcal{F}_{\text{final}} = \mathcal{F}_{\text{seg}} \;\cup\; \operatorname{Top}\!\bigl(N_{\text{sel}} - |\mathcal{F}_{\text{seg}}|,\;\mathcal{H}_t \setminus \mathcal{F}_{\text{seg}}\bigr),
\end{equation}
where $\operatorname{Top}(n, \cdot)$ selects the $n$ highest-scoring frames from the given candidate set, and $\mathcal{H}_t$ denotes the complete set of all frames in the historical frame buffer up to the current time $t$. Both cases guarantee exactly $N_{\text{sel}}$ frames while strictly preserving multi-segment diversity.

\subsection{Pose-Aware Spatial Memory}
\label{sec:spatial_compression}
\vspace{-1mm}

\begin{figure}[tb]  
    \centering          \includegraphics[width=1.0\linewidth]{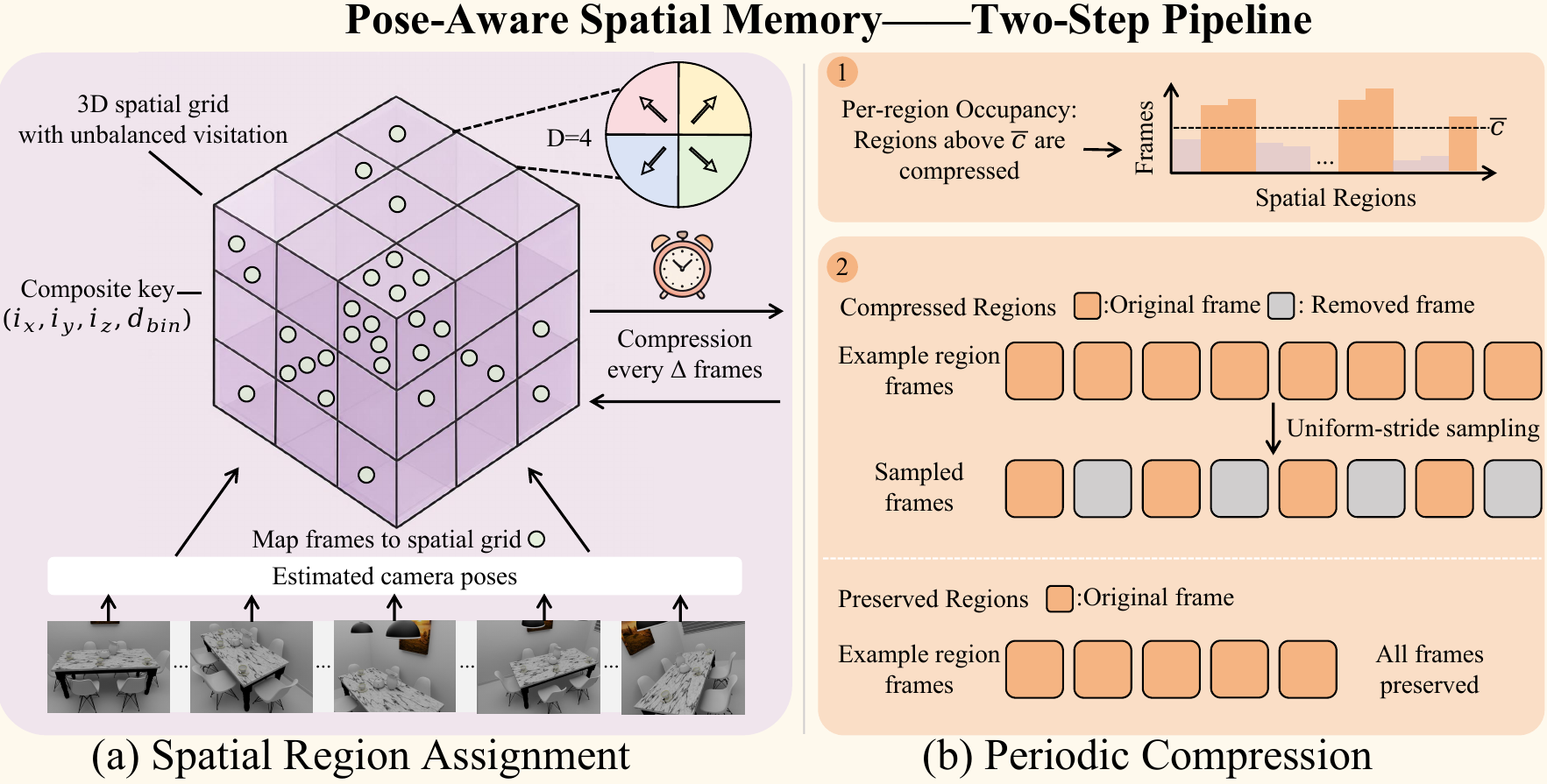}  
    \caption{%
    \textbf{Pose-Aware Spatial Memory pipeline.}
    \textbf{(a)}~Spatial Region Assignment: each frame is mapped into an isometric $K^{3}{\times}D$ grid cell indexed by its 3D position and viewing direction, with a grow-only bounding box that keeps assignments stable across deletions.
    \textbf{(b)}~Periodic Compression: regions exceeding the mean occupancy are thinned via uniform-stride sampling while anchor remains protected; deleted entries are tombstoned so metadata persists.
    }
    \label{fig:spatial_pipeline}
\end{figure}

Under certain cases, camera trajectories may revisit certain scene regions densely while leaving others sparsely covered, causing the KV cache to accumulate many redundant key frames in over-represented areas. This not only leads to unbounded growth of storage for KV states, but also dilutes the selection pool with redundant views, reducing the diversity. We therefore maintain a \textit{pose-aware spatial memory} that organizes cached KV states by utilizing their estimated camera poses and periodically compresses over-represented regions. An overview of the two-step pipeline is  in Fig.~\ref{fig:spatial_pipeline}.

\noindent\textbf{Spatial Region Assignment.}
After each frame $t$ passes through the camera prediction head, we store its 3D position $\mathbf{p}_t \in \mathbb{R}^3$ and optical-axis direction $\mathbf{d}_t \in \mathbb{R}^3$ . The scene is partitioned into a uniform $K{\times}K{\times}K$ grid based on the \textit{historical} bounding box of all observed camera positions:
\begin{equation}
  (i_x, i_y, i_z) = \left\lfloor \frac{\mathbf{p}_t - \mathbf{b}_{\min}}{g} \right\rfloor,
\end{equation}
where $\mathbf{b}_{\min}$ is the minimum corner of the bounding box and $g = \lVert \mathbf{b}_{\max} - \mathbf{b}_{\min} \rVert / K$ is the adaptive cell size derived from the scene diameter. The viewing direction is quantized into $D$ azimuth bins:
\begin{equation}
  d_{\mathrm{bin}} = \left\lfloor \frac{\mathrm{atan2}(d_z,\, d_x) + \pi}{2\pi / D} \right\rfloor.
\end{equation}
Each frame is assigned to the spatial region identified by composite key $(i_x, i_y, i_z, d_{\mathrm{bin}})$, yielding up to $K^3 \!\times\! D$ distinct regions. We set $K{=}3$ and $D{=}4$ in practice. Notably, the historical bounding box only grows and never shrinks, ensuring region assignments remain stable even after frame deletion.

\noindent\textbf{Periodic Compression.}
After processing every $\Delta$ frames, we assess the spatial distribution of frames and selectively thin over-populated regions. Let $\{R_1, \dots, R_P\}$ denote the occupied regions and $|R_p|$ the number of compressible frames in $R_p$ (anchor frame kept). The average occupancy is:
\begin{equation}
  \bar{c} = \frac{1}{P} \sum_{p=1}^{P} |R_p|.
\end{equation}
Regions with $|R_p| > \bar{c}$ are deemed over-represented. For each such region, we retain a fraction of its frames via uniform stride sampling:
\begin{equation}
  |R_p'| = \lfloor (1-\beta)\,|R_p| \rfloor,
\end{equation}
where $\beta$ is the deletion ratio (set to 0.5). $|R_p'|$ frames are evenly spaced across the temporal span to preserve spatial diversity, and the remaining $|R_p| - |R_p'|$ frames are deleted. Under-represented regions ($|R_p| \leq \bar{c}$) are left intact. Deletion is implemented via tombstoning: KV tensors are released from memory while lightweight metadata (poses, descriptors) is preserved, keeping indices stable. 

Pose-Aware Spatial Memory bounds the memory footprint of the KV cache over arbitrarily long sequences while maintaining diverse spatial coverage: redundant views in densely visited areas are thinned, whereas unique viewpoints from sparsely visited areas are preserved.

\begin{table*}[t]
\centering
\caption{\textbf{3D reconstruction results} on 7-Scenes~\cite{shotton2013scene} and NRGBD~\cite{azinovic2022neural} at sequence lengths of 200, 300, 400, and 500 frames. Best results are in \textcolor{red}{red}, second best are \textcolor{blue}{blue}. ``OOM'' indicates out-of-memory failure.}
\hspace{-2mm}
{\footnotesize
\setlength{\tabcolsep}{2pt}
\begin{tabular}{@{}l cccc cccc cccc@{}}
\toprule
\multirow{2}{*}{Method}
  & \multicolumn{4}{c}{Acc $\downarrow$}
  & \multicolumn{4}{c}{Comp $\downarrow$}
  & \multicolumn{4}{c}{NC $\uparrow$} \\
\cmidrule(lr){2-5} \cmidrule(lr){6-9} \cmidrule(lr){10-13}
  & 200 & 300 & 400 & 500
  & 200 & 300 & 400 & 500
  & 200 & 300 & 400 & 500 \\
\midrule
\multicolumn{13}{l}{7-Scenes} \\
\midrule
VGGT~\cite{wang2025vggt}                  & OOM & OOM & OOM & OOM & OOM & OOM & OOM & OOM & OOM & OOM & OOM & OOM \\
StreamVGGT~\cite{zhuo2025streaming}      & OOM & OOM & OOM & OOM & OOM & OOM & OOM & OOM & OOM & OOM & OOM & OOM \\
STream3R$\beta$~\cite{lan2025stream3r}  & OOM & OOM & OOM & OOM & OOM & OOM & OOM & OOM & OOM & OOM & OOM & OOM \\
Spann3R~\cite{wang20253d}            & 0.0403 & 0.0765 & 0.0797 & 0.0707 & \textcolor{blue}{0.0214} & 0.0278 & 0.0275 & \textcolor{red}{0.0187} & 0.5730 & 0.5464 & 0.5428 & 0.5402 \\
Point3R~\cite{wu2025point3r}            & 0.0371 & 0.0457 & 0.0518 & 0.0585 & \textcolor{red}{0.0203} & 0.0277 & 0.0242 & 0.0316 & 0.5778 & 0.5646 & 0.5604 & 0.5542 \\
CUT3R~\cite{wang2025continuous}                & 0.0894 & 0.1269 & 0.1604 & 0.1827 & 0.0483 & 0.0629 & 0.0995 & 0.0808 & 0.5642 & 0.5429 & 0.5351 & 0.5309 \\
TTT3R~\cite{chen2025ttt3r}                & \textcolor{red}{0.0275} & 0.0402 & 0.0502 & 0.0651 & 0.0232 & 0.0245 & 0.0263 & 0.0295 & 0.5804 & 0.5647 & 0.5575 & 0.5522 \\
InfiniteVGGT~\cite{yuan2026infinitevggt}  & 0.0454 & 0.0432 & 0.0421 & 0.0398 & 0.0307 & 0.0269 & 0.0269 & 0.0236 & 0.5817 & 0.5695 & 0.5649 & \textcolor{blue}{0.5617} \\
Random sampling ($N{=}48$)               & 0.0439 & 0.0433 & 0.0432 & 0.0409 & 0.0287 & 0.0237 & \textcolor{blue}{0.0237} & 0.0209 & \textcolor{blue}{0.5849} & \textcolor{red}{0.5745} & \textcolor{red}{0.5696} & 0.5614 \\
Uniform sampling ($N{=}48$)              & 0.0382 & 0.0379 & \textcolor{blue}{0.0378} & \textcolor{blue}{0.0392}    & 0.0277 & 0.0238 & 0.0240 & 0.0227    & 0.5838 & 0.5705 & 0.5659 & 0.5600    \\
Sliding window ($N{=}48$)                & 0.0327 & \textcolor{blue}{0.0340} & 0.0451 & 0.0530    & 0.0253 & \textcolor{blue}{0.0221} & 0.0244 & 0.0227    & 0.5843 & 0.5689 & 0.5644 & 0.5591    \\
RetrieveVGGT (Ours)                      & \textcolor{blue}{0.0324} & \textcolor{red}{0.0314} & \textcolor{red}{0.0316} & \textcolor{red}{0.0312} & 0.0252 & \textcolor{red}{0.0211} & \textcolor{red}{0.0224} & \textcolor{blue}{0.0199} & \textcolor{red}{0.5852} & \textcolor{blue}{0.5708} & \textcolor{blue}{0.5661} & \textcolor{red}{0.5617} \\
\midrule    
\multicolumn{13}{l}{NRGBD} \\
\midrule  
VGGT~\cite{wang2025vggt}                 & OOM & OOM & OOM & OOM & OOM & OOM & OOM & OOM & OOM & OOM & OOM & OOM \\
StreamVGGT~\cite{zhuo2025streaming}      & OOM & OOM & OOM & OOM & OOM & OOM & OOM & OOM & OOM & OOM & OOM & OOM \\
STream3R$\beta$~\cite{lan2025stream3r}  & OOM & OOM & OOM & OOM & OOM & OOM & OOM & OOM & OOM & OOM & OOM & OOM \\
Spann3R~\cite{wang20253d}            & 0.0709 & 0.1002 & 0.1417 & 0.1829 & \textcolor{blue}{0.0181} & 0.0262 & 0.0447 & 0.0487 & 0.6132 & 0.5953 & 0.5855 & 0.5794 \\
Point3R~\cite{wu2025point3r}            & 0.0636 & 0.0757 & 0.1045 & 0.1208 & 0.0191 & \textcolor{red}{0.0143} & \textcolor{blue}{0.0299} & \textcolor{blue}{0.0312} & 0.6220 & 0.6136 & 0.6067 & 0.6139 \\
CUT3R~\cite{wang2025continuous}                & 0.1362 & 0.2327 & 0.3231 & 0.3341 & 0.0299 & 0.0687 & 0.0992 & 0.1350 & 0.5987 & 0.5704 & 0.5512 &  0.5506 \\
TTT3R~\cite{chen2025ttt3r}                & 0.1298 & 0.2278 & 0.3146 & 0.3196 & 0.0304 & 0.0703 & 0.1103 & 0.1440 & 0.6014 & 0.5777 & 0.5535 & 0.5510 \\
InfiniteVGGT~\cite{yuan2026infinitevggt}  & 0.0461 & \textcolor{blue}{0.0532} & \textcolor{blue}{0.0651} & \textcolor{blue}{0.0696} & 0.0200 & 0.0244 & 0.0348 & 0.0370 & \textcolor{blue}{0.6520} & \textcolor{blue}{0.6431} & \textcolor{blue}{0.6503} & \textcolor{blue}{0.6433} \\
Random sampling ($N{=}48$)               & 0.0433 & 0.0595 & 0.0777 & 0.0790 & 0.0189 & 0.0233 & 0.0311 & 0.0312 & 0.6435 & 0.6366 & 0.6305 & 0.6296 \\
Uniform sampling ($N{=}48$)              & \textcolor{blue}{0.0422} & 0.0578 & 0.0773 & 0.0760 & 0.0202 & 0.0254 & 0.0401 & 0.0371 & 0.6468 & 0.6415 & 0.6374 & 0.6359 \\
Sliding window ($N{=}48$)                & 0.0869 & 0.1191 & 0.1501 & 0.1622    & 0.0296 & 0.0410 & 0.0626 & 0.0539    & 0.6135 & 0.6152 & 0.6028 & 0.6048    \\
RetrieveVGGT (Ours)                      & \textcolor{red}{0.0351} & \textcolor{red}{0.0454} & \textcolor{red}{0.0464} & \textcolor{red}{0.0500} & \textcolor{red}{0.0150} & \textcolor{blue}{0.0222} & \textcolor{red}{0.0280} & \textcolor{red}{0.0294} & \textcolor{red}{0.6649} & \textcolor{red}{0.6553} & \textcolor{red}{0.6566} & \textcolor{red}{0.6554} \\
\bottomrule
\end{tabular}}   
\label{tab:3d_recon}
\end{table*}
\section{Experiments}
\vspace{-2mm}
\subsection{Implementation Details}
\label{sec:impl_details}
\vspace{-1mm}

RetrieveVGGT is built upon the open-source StreamVGGT~\cite{zhuo2025streaming} checkpoint without fine-tuning. We set $N{=}48$ and evaluate on three tasks: 3D reconstruction on 7-Scenes~\cite{shotton2013scene} and NRGBD~\cite{azinovic2022neural}, video depth estimation on Bonn~\cite{palazzolo2019refusion}, and camera pose estimation on TUM-dynamics~\cite{sturm2012benchmark}. Additional short-sequence results are in Appendix D. All experiments run on a single NVIDIA RTX 5880 Ada GPU. All hyperparameters and their default values are in Appendix C.

\vspace{-2mm}
\subsection{3D Reconstruction}
\label{sec:exp_3d_recon}
\vspace{-1mm}

We evaluate 3D reconstruction on 7-Scenes~\cite{shotton2013scene} and NRGBD~\cite{azinovic2022neural} with 200--500 frames (Tab.~\ref{tab:3d_recon}). RetrieveVGGT achieves the best results across most settings, outperforming InfiniteVGGT~\cite{yuan2026infinitevggt}'s query-agnostic pruning while remaining robust where TTT3R and Point3R degrade; VGGT~\cite{wang2025vggt}, StreamVGGT~\cite{zhuo2025streaming}, and STream3R$\beta$~\cite{lan2025stream3r} encounter OOM at even 200 frames. Under the same budget $N{=}48$, random and uniform sampling lack geometric guidance and degrade severely on NRGBD (Acc error ${\approx}82\%$), while sliding window discards all earlier context (Acc 0.087--0.162). These results confirm that attention-derived relevance---not merely the budget---drives improvement. Qualitative comparisons are provided in Appendix E.

\begin{table*}[t]
\centering
\caption{\textbf{Video Depth Estimation on Bonn}. We report Abs Rel$\downarrow$ and $\delta{<}1.25$ ($\uparrow$) under per-sequence scale alignment at sequence lengths of 200--500 frames. Best in \textcolor{red}{red}, second best \textcolor{blue}{blue}. ``OOM'' indicates out-of-memory failure.}
{\small
\setlength{\tabcolsep}{6.5pt}
\begin{tabular}{@{}l cccc cccc@{}}
\toprule
\multirow{2}{*}{Method}
  & \multicolumn{4}{c}{Abs Rel $\downarrow$}
  & \multicolumn{4}{c}{$\delta{<}1.25$ $\uparrow$} \\
\cmidrule(lr){2-5} \cmidrule(lr){6-9}
  & 200 & 300 & 400 & 500
  & 200 & 300 & 400 & 500 \\
\midrule
VGGT~\cite{wang2025vggt}                  & OOM & OOM & OOM & OOM & OOM & OOM & OOM & OOM \\
StreamVGGT~\cite{zhuo2025streaming}      & OOM & OOM & OOM & OOM & OOM & OOM & OOM & OOM \\
STream3R$\beta$~\cite{lan2025stream3r} & OOM & OOM & OOM & OOM & OOM & OOM & OOM & OOM \\
Spann3R~\cite{wang20253d}            & 0.2460 & 0.2680 & 0.2746 & 0.2608 & 0.6802 & 0.6574 & 0.6393 & 0.6476 \\
Point3R~\cite{wu2025point3r}            & 0.0709 & 0.0829 & 0.0826 & 0.0802 & 0.9555 & 0.9475 & 0.9468 & 0.9490 \\
CUT3R~\cite{wang2025continuous}                & 0.0726 & 0.0888 & 0.0898 & 0.0847 & 0.9469 & 0.9387 & 0.9338 & 0.9387 \\
TTT3R~\cite{chen2025ttt3r}                & 0.0681 & 0.0789 & 0.0779 & 0.0755 & 0.9531 & 0.9499 & 0.9512 & 0.9534 \\
InfiniteVGGT~\cite{yuan2026infinitevggt}  & \textcolor{blue}{0.0637} & \textcolor{blue}{0.0730} & \textcolor{blue}{0.0717} & \textcolor{blue}{0.0699} & \textcolor{blue}{0.9659} & \textcolor{blue}{0.9574} & \textcolor{blue}{0.9585} & \textcolor{blue}{0.9600} \\
RetrieveVGGT (Ours)                      & \textcolor{red}{0.0604} & \textcolor{red}{0.0698} & \textcolor{red}{0.0699} & \textcolor{red}{0.0669} & \textcolor{red}{0.9677} & \textcolor{red}{0.9620} & \textcolor{red}{0.9626} & \textcolor{red}{0.9634} \\
\bottomrule
\end{tabular}}
\label{tab:depth}
\vspace{-2mm}
\end{table*}

\vspace{-2mm}
\subsection{Video Depth Estimation}
\label{sec:exp_depth}
\vspace{-1mm}

Beyond scene-level 3D reconstruction, we further evaluate per-frame depth accuracy and inter-frame depth consistency, which are crucial indicators of whether a streaming method maintains coherent geometry over time. We adopt Bonn~\cite{palazzolo2019refusion} as the benchmark, since it provides the longest available continuous RGBD sequences among existing indoor datasets, following existing protocols~\cite{chen2025ttt3r,yuan2026infinitevggt}. Specifically, we select continuous sequences of 200 to 500 frames (beginning after the initial 30 frames) from five scenes, and report Absolute Relative Error (Abs Rel$\downarrow$) and $\delta{<}1.25$ ($\uparrow$) under per-sequence scale alignment. As in Tab.~\ref{tab:depth}, RetrieveVGGT achieves the best depth estimation accuracy across all sequence lengths on both metrics, consistently outperforming the second-best InfiniteVGGT.

\begin{figure}[tb]  
    \centering      
    \includegraphics[width=1.0\linewidth]{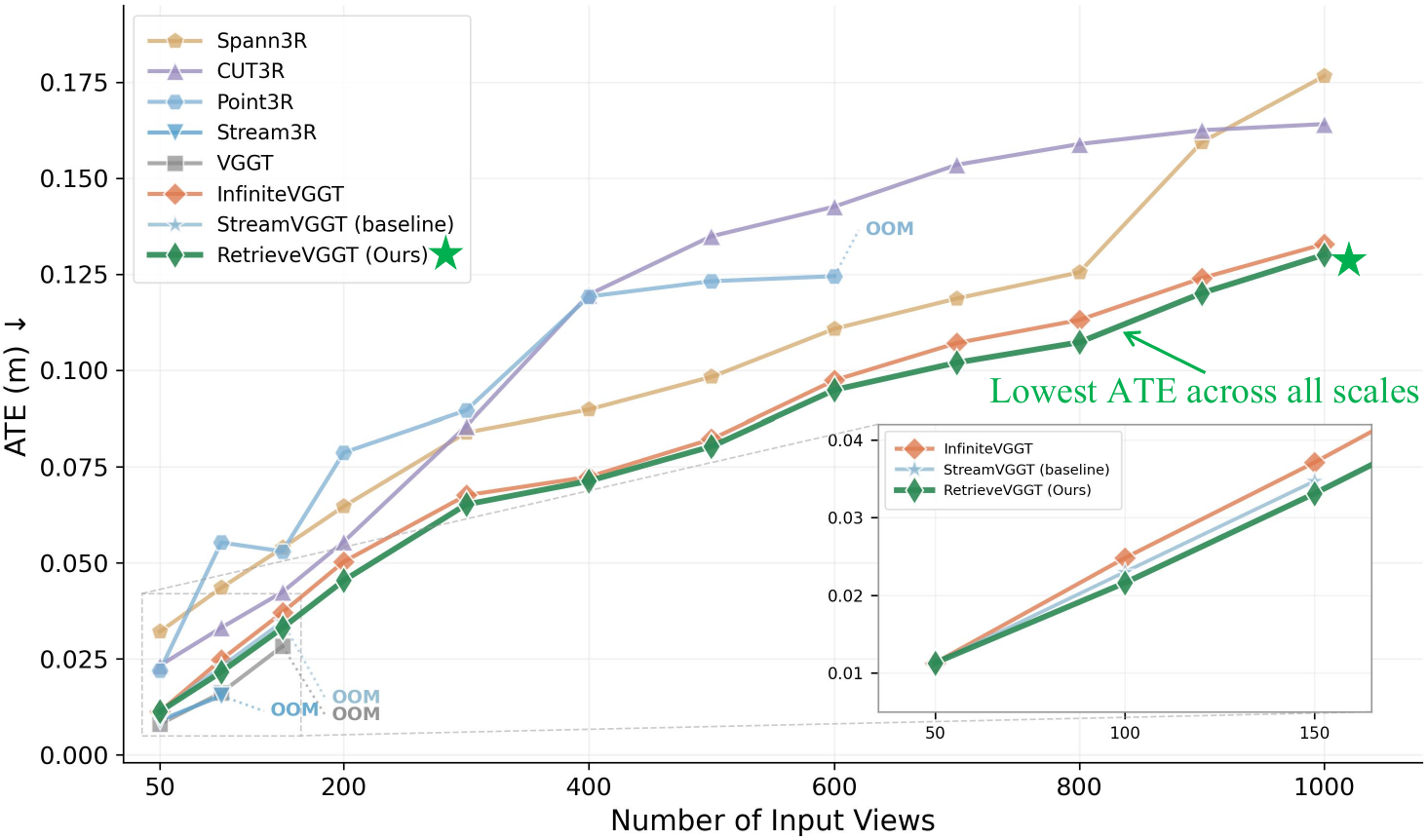}
    \caption{\textbf{Influence of input length} on ATE on TUM-dynamics (50--1000 frames). Inset: short-sequence regime ($\leq$150 frames). OOM runs truncated at their last successful length.}
    \label{fig:pose_curve}

\end{figure}
\vspace{-4mm}

\subsection{Camera Pose Estimation}
\label{sec:exp_pose}
\vspace{-1mm}

We evaluate camera pose estimation on TUM-dynamics~\cite{sturm2012benchmark}. We select continuous sequences with stride~1 and vary the length from 50 to 1\,000 frames. Predicted trajectories are aligned to ground truth via Sim(3) Umeyama alignment. Fig.~\ref{fig:pose_curve} plots ATE as a function of sequence length. RetrieveVGGT achieves the lowest error almost across all lengths, even on the very short input length where StreamVGGT is not OOM.

\vspace{-2mm}

\section{Ablation Studies}
\label{sec:ablation}

\begin{table*}[t]
\footnotesize
\centering
\caption{\textbf{Ablation Studies} on 7-Scenes (500 frames).
\textbf{(a)}~Descriptor feature for frame selection.
\textbf{(b)}~Selection strategy and threshold.
\textbf{(c)}~Frame budget $N$ (1 anchor + $N{-}1$ selected).
\textbf{(d)}~Compression strategy (both compressed variants retain about 320 active keyframes).
\textbf{(e)}~Spatial grid resolution.}
\subcaptionbox{Descriptor Feature.\label{tab:ablation_descriptor}}{
    \setlength{\tabcolsep}{2.5mm}
    \begin{tabular}{@{}l cc@{}}
    \toprule
    Descriptor Source & Acc $\downarrow$ & Comp $\downarrow$ \\
    \midrule
    DINOv2 (Layer 8, shallow) & 0.0740 & 0.0251 \\
    DINOv2 (Layer 12, middle) & 0.0760 & 0.0244 \\
    DINOv2 (Layer 20, deep) & 0.0865 & 0.0291 \\
    First-Layer Selection (Ours) & \textbf{0.0312} & \textbf{0.0199} \\
    Every-Layer Selection & \underline{0.0325} & \underline{0.0217} \\
    \bottomrule
    \end{tabular}
}\hspace{3mm}
\subcaptionbox{Selection Strategy \& Threshold.\label{tab:ablation_segment}}{
    \setlength{\tabcolsep}{1.5mm}
    \begin{tabular}{@{}l cc@{}}
    \toprule
    Selection Strategy & Acc $\downarrow$ & Comp $\downarrow$ \\
    \midrule
    Top-$K$ (no Segment Sampling) & 0.0530 & 0.0227 \\
    Probabilistic Sampling & 0.0341 & \textbf{0.0196} \\
    Segment Sampling ($\tau = \mu$) & \underline{0.0335} & 0.0210 \\
    Segment Sampling ($\tau = \mu{+}0.3\sigma$) & \textbf{0.0312} & \underline{0.0199} \\
    Segment Sampling ($\tau = \mu{+}0.6\sigma$) & 0.0381 & 0.0208 \\
    \bottomrule
    \end{tabular}
}\vspace{1mm}\\
\hspace{-1mm}
\subcaptionbox{Frame Budget.\label{tab:ablation_budget}}{
    \setlength{\tabcolsep}{1.2mm}
    \begin{tabular}{@{}c cc@{}}
    \toprule
    Budget $N$ & Acc $\downarrow$ & Comp $\downarrow$ \\
    \midrule
    12 & 0.0488 & 0.0199 \\
    24 & \underline{0.0360} & \textbf{0.0188} \\
    48 (Ours) & \textbf{0.0312} & \underline{0.0199} \\
    \bottomrule
    \end{tabular}
}\hspace{2mm}
\subcaptionbox{Compression Strategy.\label{tab:ablation_compression}}{
    \setlength{\tabcolsep}{1.2mm}
    \begin{tabular}{@{}l cc@{}}
    \toprule
    Strategy & Acc $\downarrow$ & Comp $\downarrow$ \\
    \midrule
    None (full cache) & \textbf{0.0312} & \textbf{0.0199} \\
    Random deletion &0.0376  &0.0228  \\
    Pose-aware (Ours) & \underline{0.0327} & \underline{0.0206} \\
    \bottomrule
    \end{tabular}
}\hspace{2mm}
\subcaptionbox{Grid Resolution.\label{tab:ablation_grid}}{
    \setlength{\tabcolsep}{1.2mm}
    \begin{tabular}{@{}l cc@{}}
    \toprule
    Configuration & Acc $\downarrow$ & Comp $\downarrow$ \\
    \midrule
    w/o position &0.0355 &0.0216 \\
    w/o direction &\underline{0.0336} &\underline{0.0214} \\
    Full (Ours) & \textbf{0.0327} & \textbf{0.0206} \\
    \bottomrule
    \end{tabular}
}

\label{tab:ablation}

\end{table*}

\subsection{Descriptor Feature for Frame Selection}

A central design choice is which feature to use for relevance scoring. 
We compare frozen DINOv2 features at multiple depths (layers 8, 12, and 20) against features extracted from VGGT's global attention blocks. 
As in Tab.~\ref{tab:ablation_descriptor}, first global attention is significantly better, possibly due to \textbf{the VGGT pretraining already enables the global attention to find out relevant areas for SfM}. 
Furthermore, performing independent selection at every layer yields no gain over the adopted first-layer-only selection.

\subsection{Selection Strategy: Top-\texorpdfstring{$K$}{K} vs.\ Segment Sampling}

We evaluate the impact of Segment Sampling and the sensitivity of threshold $\tau$ across four configurations (Tab.~\ref{tab:ablation_segment}). 
Probabilistic Sampling improves over na\"ive top-$K$ by introducing stochasticity, yet it still lacks explicit control over multi-peak coverage. 
Segment Sampling outperforms both by guaranteeing that frames are drawn from multiple relevance peaks, thereby ensuring diversity of information. 
Different $w_\textbf{thre}$ leads to some variance, but the overall performance is stable.

\subsection{Frame Budget}

We ablate the frame budget $N$ (1 anchor + $N{-}1$ selected) on 7-Scenes at 500 frames. 
As in Tab.~\ref{tab:ablation_budget}, Accuracy improves steadily with $N$ while Completeness remains stable, confirming that RetrieveVGGT effectively utilizes additional capacity. 
Notably, even at $N{=}24$, RetrieveVGGT already surpasses InfiniteVGGT's Accuracy ($0.0360$ vs.\ $0.0398$), demonstrating the efficiency of our selection strategy. 
We set $N{=}48$, the same as InfiniteVGGT for fair comparisons.

\subsection{Pose-Aware Spatial Compression}

We ablate the periodic compression module that bounds the KV cache size for long sequences. 
Without compression, all KV states accumulate indefinitely. 
Our \textit{pose-aware} compression groups cached frames by spatial region and preferentially thins over-represented regions while preserving unique viewpoints. 
For a fair comparison under the same memory budget, we also evaluate a \textit{random} baseline that deletes the same number of frames at each trigger but without considering their spatial distribution. 
Both compressed variants reduce the active KV cache from 500 to about 320 keyframes on average. 
As shown in Tab.~\ref{tab:ablation_compression}, pose-aware compression maintains reconstruction quality close to the uncompressed upper bound, whereas random deletion degrades quality due to the indiscriminate removal of informative frames.

\subsection{Spatial Grid Resolution}

We ablate the grid resolution $(K, D)$ for pose-aware compression on 7-Scenes at 500 frames, where $K$ denotes the number of position grid divisions per axis and $D$ the number of azimuth direction bins. 
As in Tab.~\ref{tab:ablation_grid}, removing either position partitioning $(K{=}1, D{=}4)$ or direction partitioning $(K{=}3, D{=}1)$ degrades reconstruction quality, confirming that both spatial and directional awareness contribute to identifying truly redundant frames. 
We adopt $(K{=}3, D{=}4)$ as the default.

\vspace{-3mm}

\section{Conclusion}
\label{sec:conclusion}
\vspace{-2mm}
We present RetrieveVGGT, a training-free streaming 3D reconstruction framework that resolves the tension between bounded memory and faithful long-sequence reconstruction. Our core insight is that each frame need only attend to a compact, query-relevant subset of history rather than the full accumulated context. We introduce Segment Sampling to ensure diversity and Pose-Aware Spatial Memory to reduce redundancy. Extensive experiments show that RetrieveVGGT achieves state-of-the-art performance while maintaining constant memory cost regardless of sequence length. Treating streaming 3D reconstruction similar to a recommender system opens a new way of tackling long context.

\clearpage
\beginappendix
\section{Retrieved Keyframe Visualization}
\label{sec:frame_retrieval_viz}

\begin{figure*}[h]
\centering
\setlength{\tabcolsep}{1.5pt}
\renewcommand{\arraystretch}{0.8}

\textbf{(a)} \texttt{complete\_kitchen} \hfill {\small \fcolorbox{red}{white}{\scriptsize\strut\,Query\,} $\rightarrow$ \fcolorbox{blue!70}{white}{\scriptsize\strut\,Retrieved Keyframes\,}}\\[2pt]
\begin{tabular}{c@{\hspace{2pt}}c@{\hspace{2pt}}c@{\hspace{2pt}}c@{\hspace{2pt}}c}
\queryimg{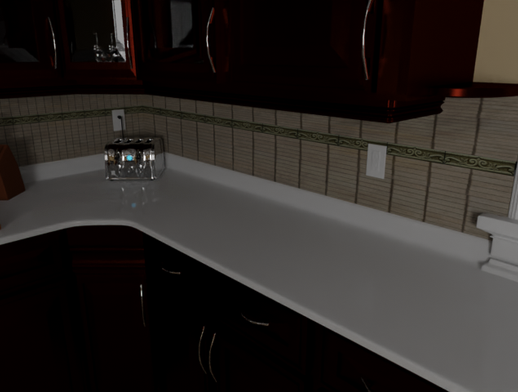} &
\retimg{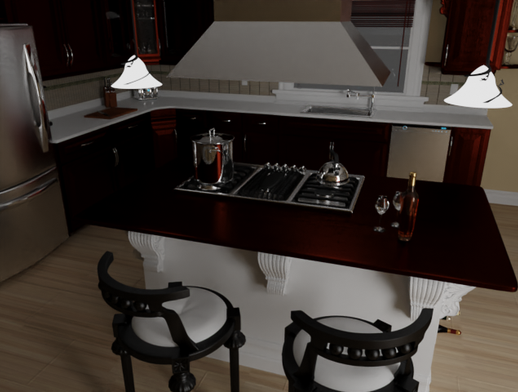} &
\retimg{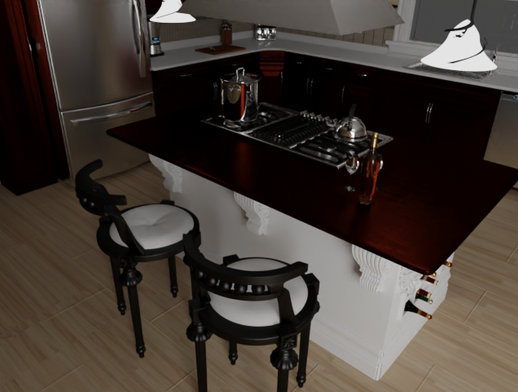} &
\retimg{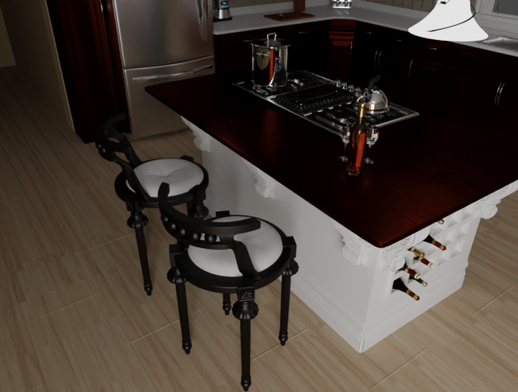} &
\retimg{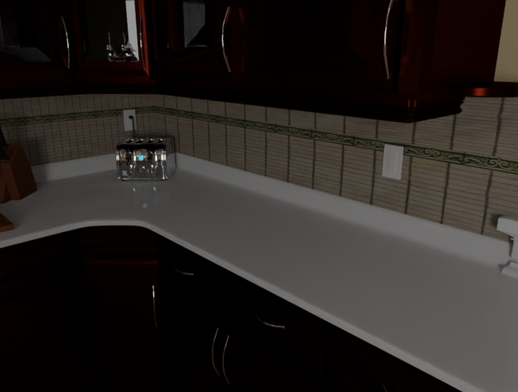} \\[-1pt]
{\scriptsize Frame 300} & \multicolumn{4}{c}{} \\[3pt]
\queryimg{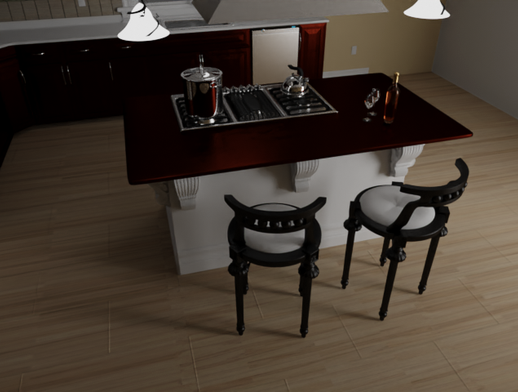} &
\retimg{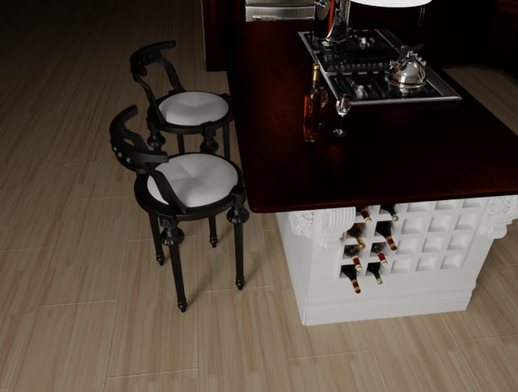} &
\retimg{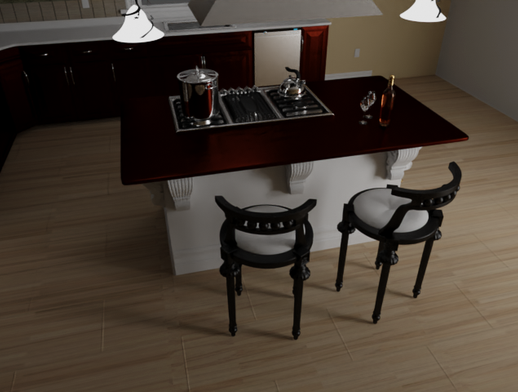} &
\retimg{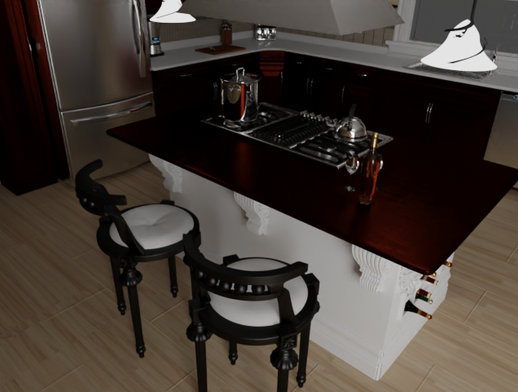} &
\retimg{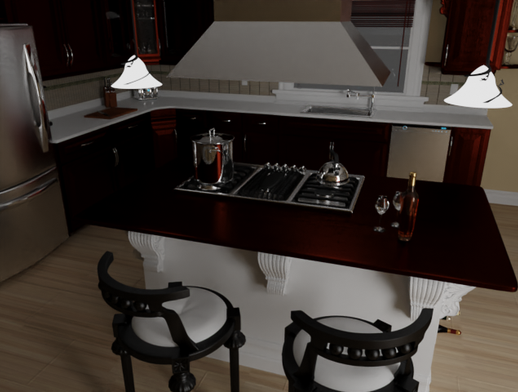} \\[-1pt]
{\scriptsize Frame 500} & \multicolumn{4}{c}{} \\
\end{tabular}

\textbf{(b)} \texttt{grey\_white\_room}\\[2pt]
\begin{tabular}{c@{\hspace{2pt}}c@{\hspace{2pt}}c@{\hspace{2pt}}c@{\hspace{2pt}}c}
\queryimg{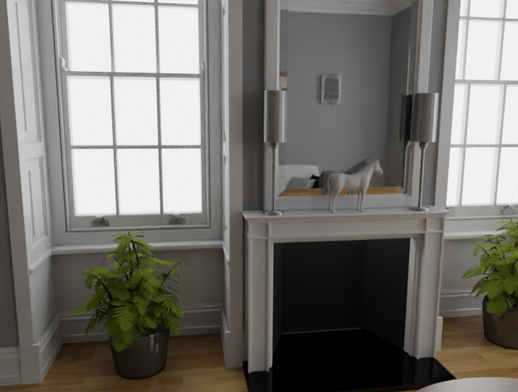} &
\retimg{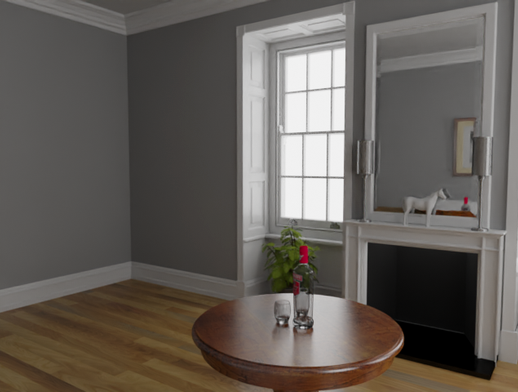} &
\retimg{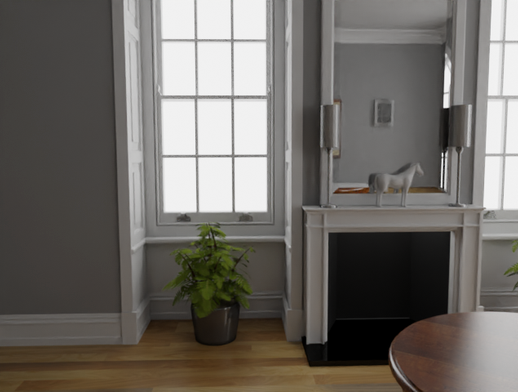} &
\retimg{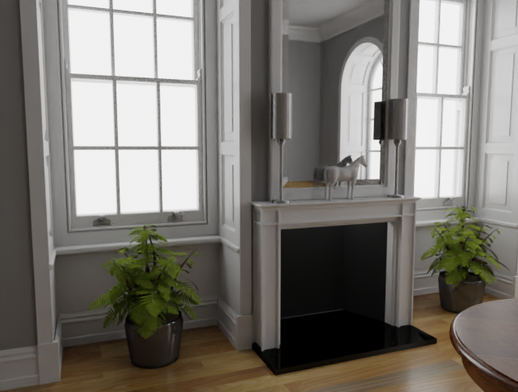} &
\retimg{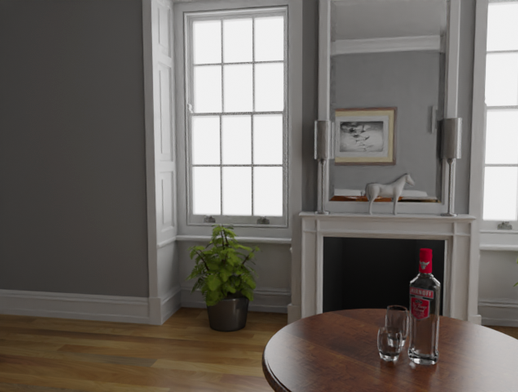} \\[-1pt]
{\scriptsize Frame 300} & \multicolumn{4}{c}{} \\[3pt]
\queryimg{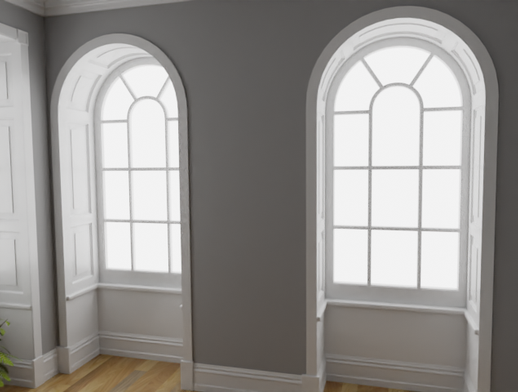} &
\retimg{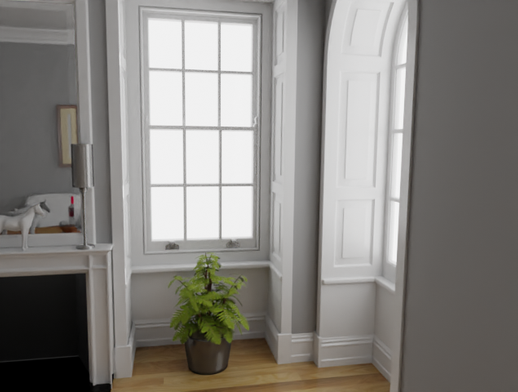} &
\retimg{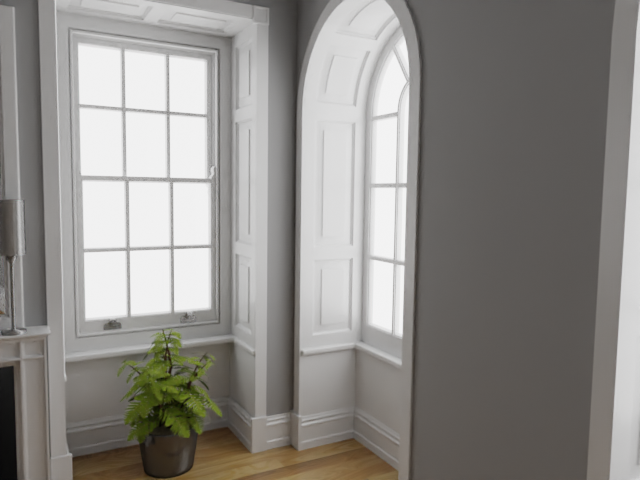} &
\retimg{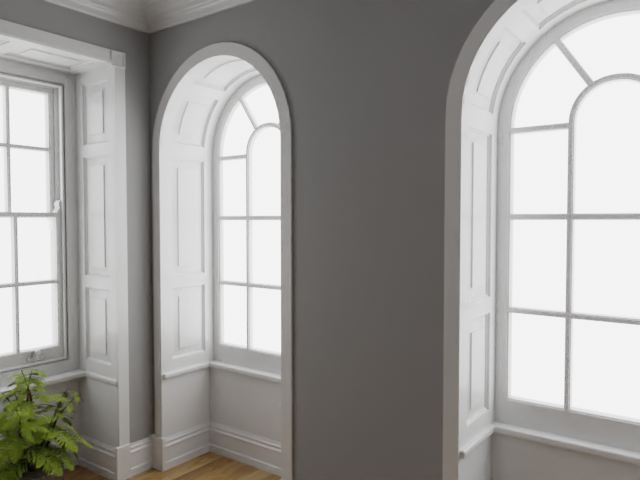} &
\retimg{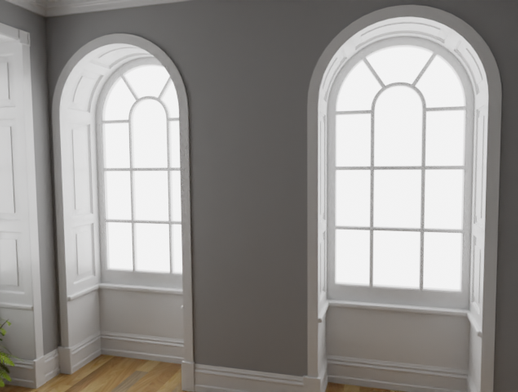} \\[-1pt]
{\scriptsize Frame 500} & \multicolumn{4}{c}{} \\
\end{tabular}
\vspace{-4pt}
\caption{\textbf{Retrieved keyframes on NRGBD~\cite{azinovic2022neural}.} The retrieved keyframes capture views of the same spatial region as the query and span a wide temporal range, confirming that Segment Sampling prevents clustering around a single peak.}
\label{fig:retrieval_nrgbd}
\end{figure*}

\begin{figure*}[!t]
\centering
\setlength{\tabcolsep}{1.5pt}
\renewcommand{\arraystretch}{0.8}

\textbf{(a)} \texttt{office/seq-02} \hfill {\small \fcolorbox{red}{white}{\scriptsize\strut\,Query\,} $\rightarrow$ \fcolorbox{blue!70}{white}{\scriptsize\strut\,Retrieved Keyframes\,}}\\[2pt]
\begin{tabular}{c@{\hspace{2pt}}c@{\hspace{2pt}}c@{\hspace{2pt}}c@{\hspace{2pt}}c}
\queryimg{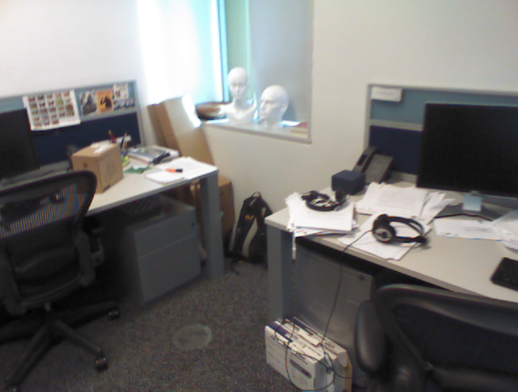} &
\retimg{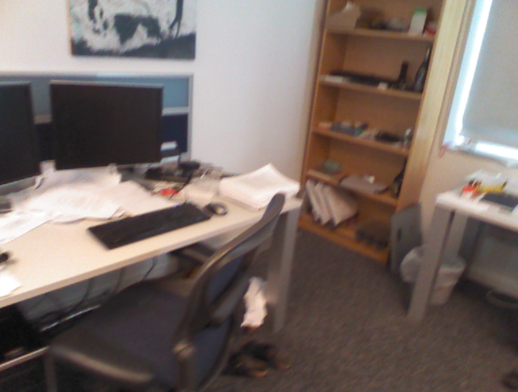} &
\retimg{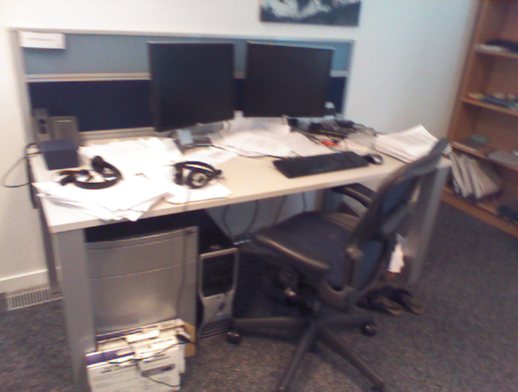} &
\retimg{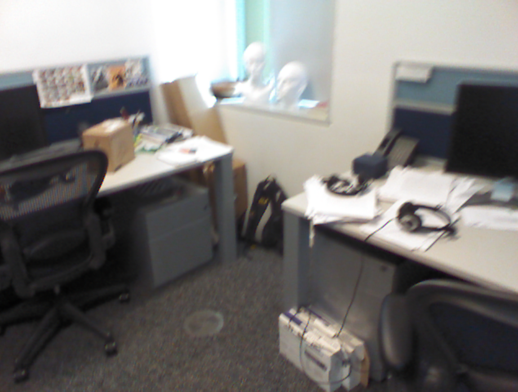} &
\retimg{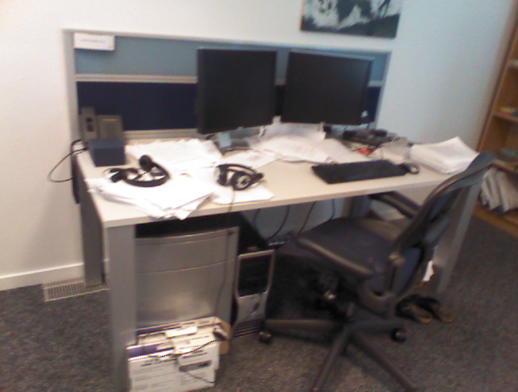} \\[-1pt]
{\scriptsize Frame 300} & \multicolumn{4}{c}{} \\[3pt]
\queryimg{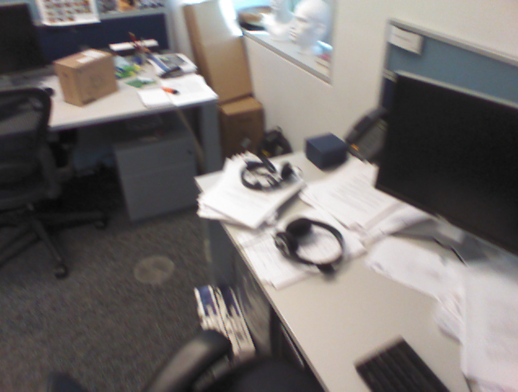} &
\retimg{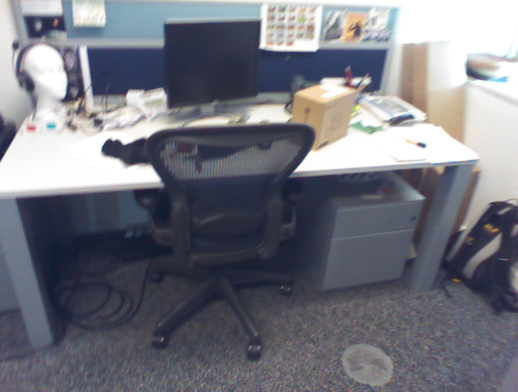} &
\retimg{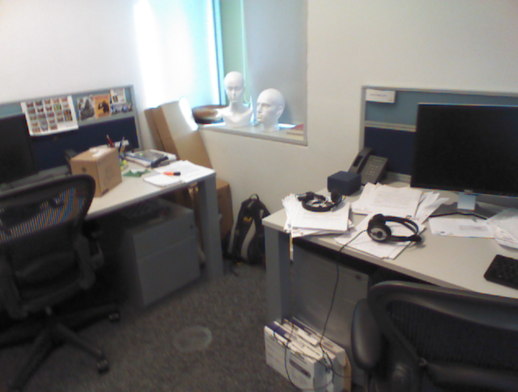} &
\retimg{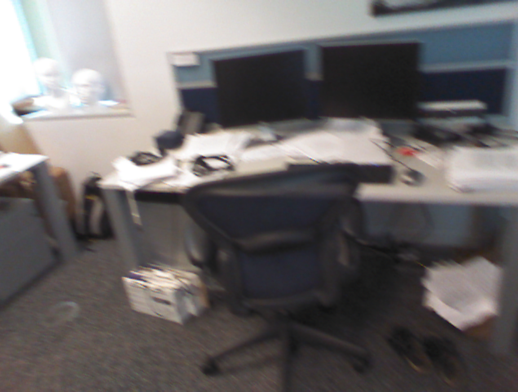} &
\retimg{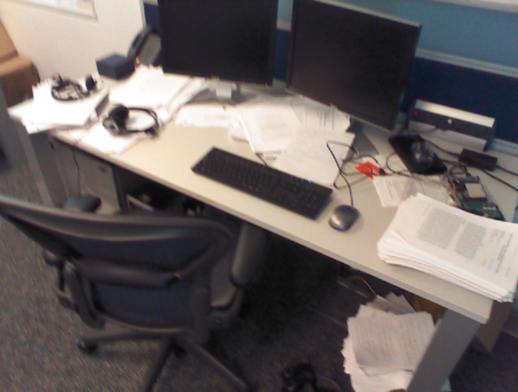} \\[-1pt]
{\scriptsize Frame 500} & \multicolumn{4}{c}{} \\
\end{tabular}

\textbf{(b)} \texttt{pumpkin/seq-07}\\[2pt]
\begin{tabular}{c@{\hspace{2pt}}c@{\hspace{2pt}}c@{\hspace{2pt}}c@{\hspace{2pt}}c}
\queryimg{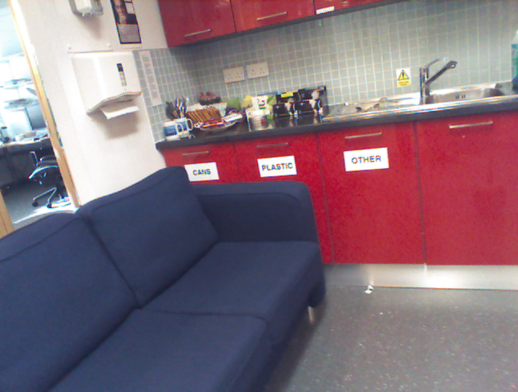} &
\retimg{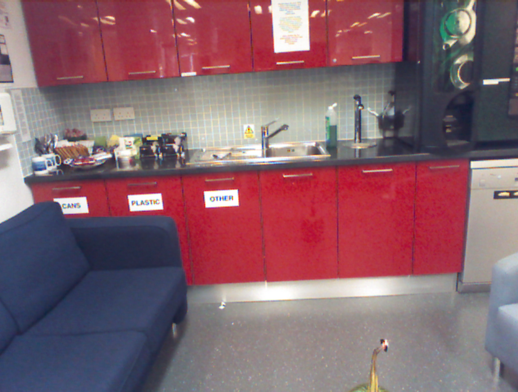} &
\retimg{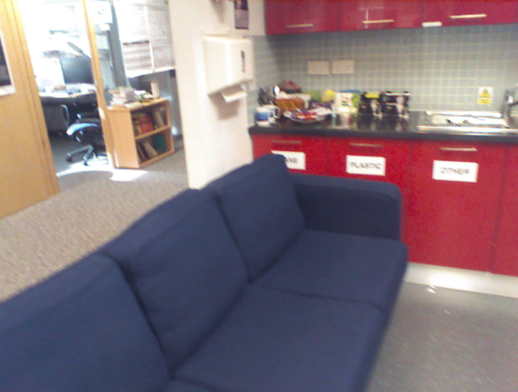} &
\retimg{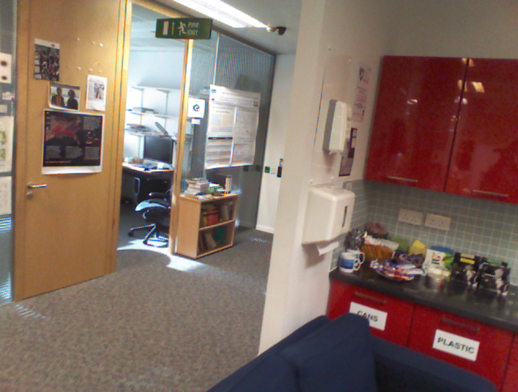} &
\retimg{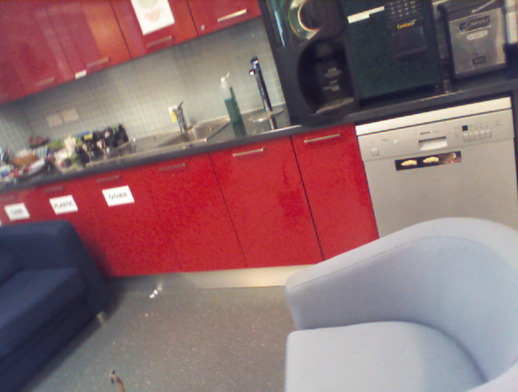} \\[-1pt]
{\scriptsize Frame 300} & \multicolumn{4}{c}{} \\[3pt]
\queryimg{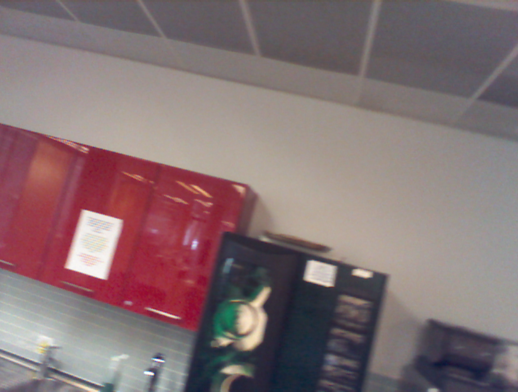} &
\retimg{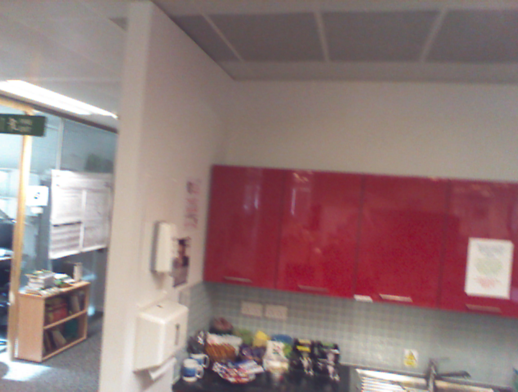} &
\retimg{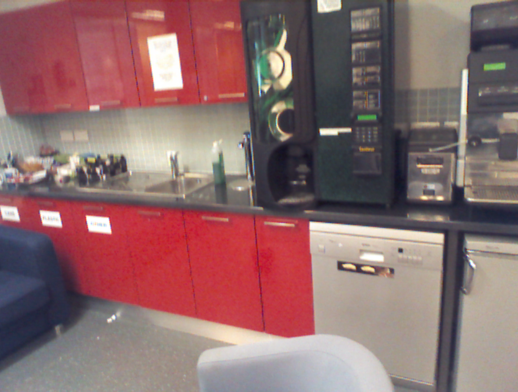} &
\retimg{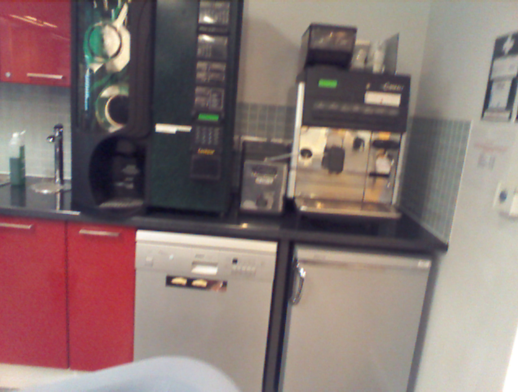} &
\retimg{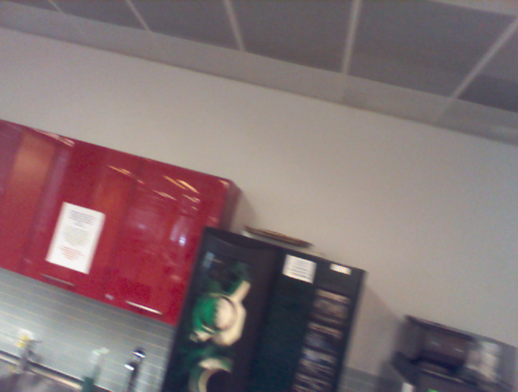} \\[-1pt]
{\scriptsize Frame 500} & \multicolumn{4}{c}{} \\
\end{tabular}

\textbf{(c)} \texttt{redkitchen/seq-12}\\[2pt]
\begin{tabular}{c@{\hspace{2pt}}c@{\hspace{2pt}}c@{\hspace{2pt}}c@{\hspace{2pt}}c}
\queryimg{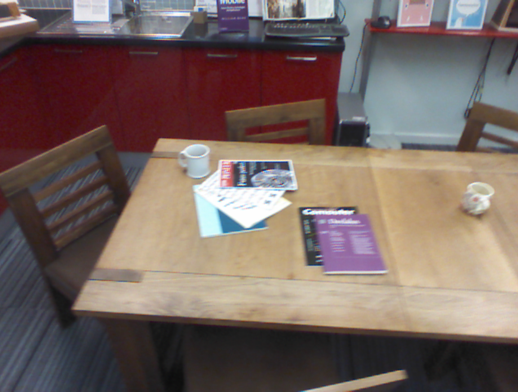} &
\retimg{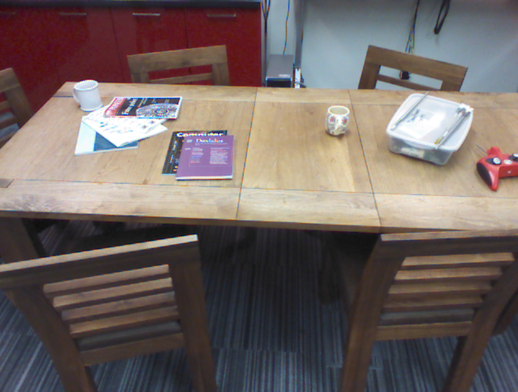} &
\retimg{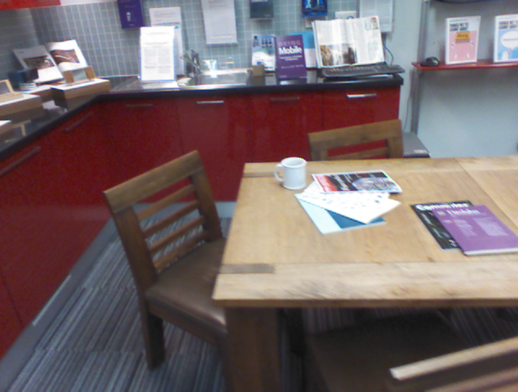} &
\retimg{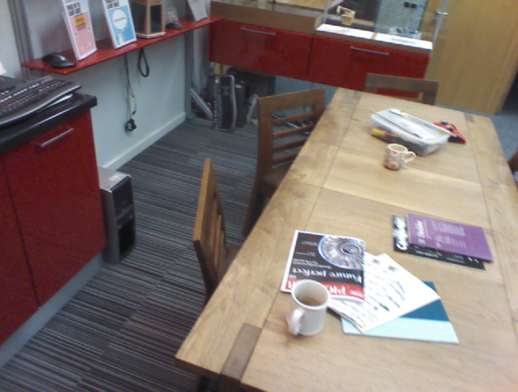} &
\retimg{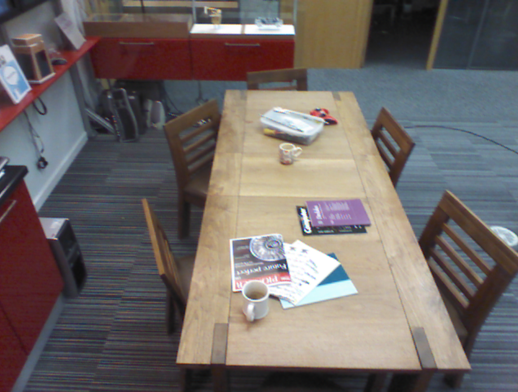} \\[-1pt]
{\scriptsize Frame 300} & \multicolumn{4}{c}{} \\[3pt]
\queryimg{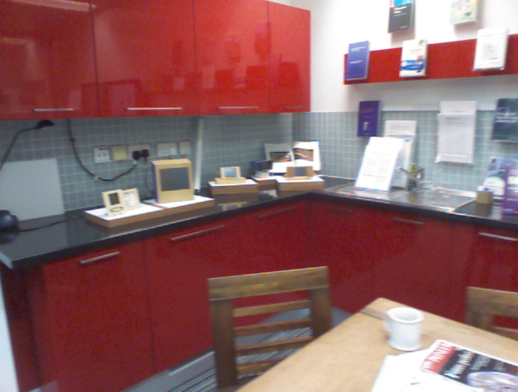} &
\retimg{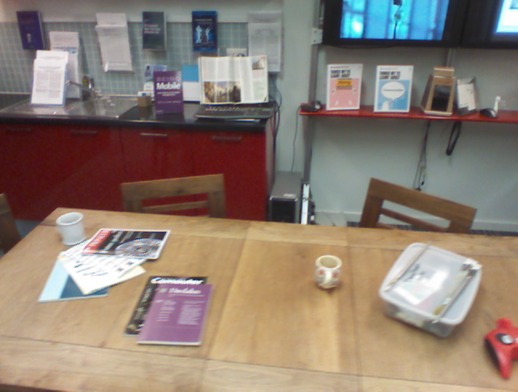} &
\retimg{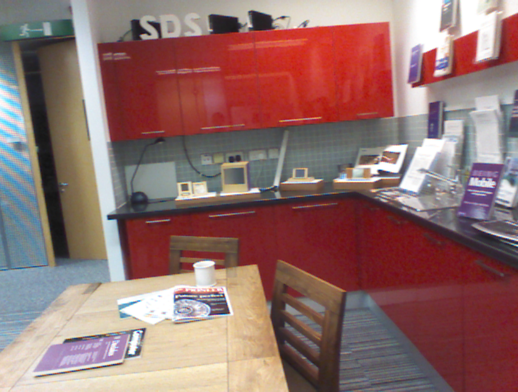} &
\retimg{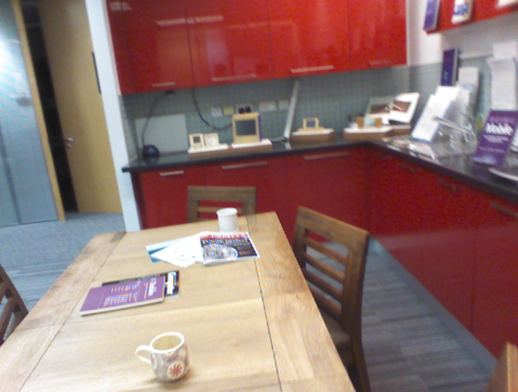} &
\retimg{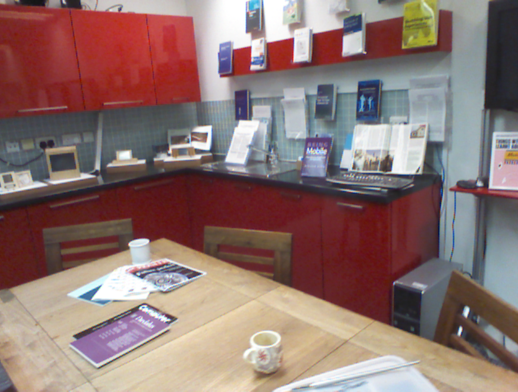} \\[-1pt]
{\scriptsize Frame 500} & \multicolumn{4}{c}{} \\
\end{tabular}
\vspace{-4pt}
\caption{\textbf{Retrieved keyframes on 7-Scenes~\cite{shotton2013scene}.} Same layout as Fig.~\ref{fig:retrieval_nrgbd}. The retrieved frames consistently cover geometrically relevant views from diverse temporal locations.}
\label{fig:retrieval_7scenes}
\end{figure*}

We visualize keyframes retrieved by RetrieveVGGT on five NRGBD~\cite{azinovic2022neural} (Fig.~\ref{fig:retrieval_nrgbd}) and 7-Scenes~\cite{shotton2013scene} (Fig.~\ref{fig:retrieval_7scenes}) scenes. For each, we show the query frame (\textcolor{red}{red border}) at Frames~300 and 500 with four retrieved keyframes (\textcolor{blue!70}{blue border}) selected via Segment Sampling from the \textit{entire} history.

\noindent\textbf{Analysis.}
\textbf{(1)~Query-adaptive:} when the query observes a different region at Frame~300 vs.\ 500, retrieved sets change accordingly---each frame selects its own context rather than sharing a globally pruned cache~\cite{yuan2026infinitevggt}.
\textbf{(2)~Geometric relevance:} retrieved keyframes depict the same or nearby regions as the query, even when separated by hundreds of frames, confirming that Q-K similarity at the first global attention layer encodes geometric correspondence without any extra learned module.
\textbf{(3)~Temporal diversity:} Segment Sampling distributes retrieved keyframes across distinct temporal segments, providing complementary multi-view coverage rather than clustering around a single peak.

\begin{table}[t]
    \centering
    \caption{\textbf{Ablation on relevance scoring functions} on 7-Scenes (500 frames, budget $N{=}48$). Raw dot product achieves the best Accuracy, confirming that magnitude information in Q-K descriptors carries important geometric signals.}
    \label{tab:ablation_sim}
    \begin{tabular}{lccc}
        \toprule
        Scoring Function & Acc $\downarrow$ & Comp $\downarrow$ & NC $\uparrow$ \\
        \midrule
        Negative L2 Distance & 0.0576 & 0.0227 & 0.5665 \\
        Cosine Similarity & \underline{0.0330} & \underline{0.0209} & \underline{0.5715} \\
        \rowcolor{gray!15} Raw Dot Product (Ours) & \textbf{0.0314} & \textbf{0.0199} & \textbf{0.5720} \\
        \bottomrule
    \end{tabular}
\end{table}

\section{Ablation on Relevance Scoring Functions}
\label{sec:ablation_sim}

Recent LLM KV cache management methods---Quest~\cite{tang2024quest}, H2O~\cite{zhang2023h2o}, and SnapKV~\cite{li2024snapkv}---show that raw pre-softmax Q$\cdot$K scores suffice to identify important KV entries. Following this principle, RetrieveVGGT uses the un-normalized dot product $r(t, i) = \frac{1}{H}\sum_{h=1}^{H} \langle \bar{\mathbf{q}}_t[h,:],\; \bar{\mathbf{k}}_{i}[h,:] \rangle$ as the frame-level relevance score, which jointly considers direction and magnitude of the Q-K descriptors. We ablate this choice against two alternatives that differ in how they treat magnitude:
\begin{itemize}[nosep,leftmargin=*]
    \item \textbf{Raw Dot Product} (Ours): $s = \frac{1}{H}\sum_{h=1}^{H} \langle \bar{\mathbf{q}}_t[h,:],\; \bar{\mathbf{k}}_{i}[h,:] \rangle$. Preserves both direction and magnitude of the learned Q-K descriptors.
    \item \textbf{Cosine Similarity}: $s = \frac{\sum_{h=1}^{H} \langle \bar{\mathbf{q}}_t[h,:],\; \bar{\mathbf{k}}_{i}[h,:] \rangle}{\sqrt{\sum_{h=1}^{H} \|\bar{\mathbf{q}}_t[h,:]\|^2}\,\sqrt{\sum_{h=1}^{H} \|\bar{\mathbf{k}}_{i}[h,:]\|^2}}$. Discards magnitude, retaining only directional agreement.
    \item \textbf{Negative L2 Distance}: $s = -\sqrt{\frac{1}{H}\sum_{h=1}^{H} \|\bar{\mathbf{q}}_t[h,:] - \bar{\mathbf{k}}_{i}[h,:]\|_2^2}$. Penalizes both directional misalignment and magnitude discrepancy.
\end{itemize}
\noindent The standard scaling factor $1/\sqrt{d_h}$~\cite{vaswani2017attention} is frame-independent and preserves relative rankings, so scaled dot product yields identical selections and is excluded.

\noindent\textbf{Analysis} (Tab.~\ref{tab:ablation_sim}).
\textbf{(1)~Magnitude encodes geometric importance.}
VGGT's key descriptors encode both viewing direction and geometric informativeness via magnitude. The raw dot product preserves both, achieving the best Accuracy.
\textbf{(2)~Cosine similarity collapses discriminability.}
After normalization, only direction remains---but pairwise cosine is ${\approx}0.98$ (Fig.~3(a), main paper), so all frames score nearly identically.
\textbf{(3)~Negative L2 penalizes informative frames.}
Since $\sum_{h=1}^{H}\|\bar{\mathbf{q}}[h,:]{-}\bar{\mathbf{k}}[h,:]\|_2^2 {=} \sum_{h=1}^{H}\|\bar{\mathbf{q}}[h,:]\|^2 {+} \sum_{h=1}^{H}\|\bar{\mathbf{k}}[h,:]\|^2 {-} 2\sum_{h=1}^{H}\langle\bar{\mathbf{q}}[h,:],\bar{\mathbf{k}}[h,:]\rangle$, high-norm keys (geometrically richer) incur \emph{larger} distances and lower scores---opposite to the dot product, which rewards them.

\section{Hyperparameters}
\label{sec:hyperparams}
Table~\ref{tab:hyperparams} lists the default hyperparameters. The frame budget $N$ controls the number of retrieved keyframes; $w_{\text{thre}}$ sets the relevance cutoff for segment identification; $\delta$ filters brief dips between segments; $\Delta$ triggers periodic spatial memory thinning; and $\beta$ controls the fraction of frames removed in over-populated regions.

\begin{table}[h]
\centering
\caption{\textbf{Hyperparameters of RetrieveVGGT.}}
\label{tab:hyperparams}
\begin{tabular}{@{}lll@{}}
\toprule
Parameter & Symbol & Value \\
\midrule
Frame budget & $N$ & 48 \\
Segment threshold coefficient & $w_{\text{thre}}$ & 0.3 \\
Segment merge gap & $\delta$ & 3 frames \\
Compression trigger interval & $\Delta$ & 200 frames \\
Deletion ratio & $\beta$ & 0.5 \\
Spatial grid resolution & $(K, D)$ & $(3, 4)$ \\
\bottomrule
\end{tabular}
\end{table}

\section{Additional Quantitative Results}
\label{sec:more_results}

The main paper reports long-sequence results (200--1000 frames), where VGGT, StreamVGGT, and STream3R$\beta$ encounter out-of-memory failures. Here we complement the above with short-sequence evaluation, enabling direct comparison with these methods. Notably, as a \textbf{training-free} method, RetrieveVGGT matches or surpasses StreamVGGT~\cite{zhuo2025streaming} even on short sequences where StreamVGGT retains the full uncompressed KV cache.

\subsection{Camera Pose Estimation}
\label{sec:more_pose}

We evaluate camera pose estimation on short sequences following the protocol of TTT3R~\cite{chen2025ttt3r}: Sintel~\cite{butler2012naturalistic} (50 frames) and TUM-dynamics~\cite{sturm2012benchmark} (90 frames). Predicted trajectories are aligned to ground truth via Sim(3) Umeyama alignment. Results are in Tab.~\ref{tab:supp_pose}.

\subsection{Video Depth Estimation}
\label{sec:more_depth}

We evaluate video depth estimation on short sequences following the protocol of TTT3R~\cite{chen2025ttt3r}: Sintel~\cite{butler2012naturalistic} (50 frames), Bonn~\cite{palazzolo2019refusion} (110 frames), and KITTI~\cite{geiger2013vision} (110 frames). We report scale-invariant Abs Rel$\downarrow$ and $\delta{<}1.25$ ($\uparrow$) under per-sequence scale alignment. Results are in Tab.~\ref{tab:supp_depth}.

\begin{table*}[!t]
    \centering
    \caption{\textbf{Camera pose estimation on short sequences.} We evaluate on Sintel~\cite{butler2012naturalistic} (50 frames) and TUM-dynamics~\cite{sturm2012benchmark} (90 frames). Best results are in \textbf{bold}, second best are \underline{underlined}. $^\dagger$Base model of RetrieveVGGT.}
    \label{tab:supp_pose}
    \resizebox{\textwidth}{!}{%
    \begin{tabular}{lc ccc ccc}
    \toprule
    \multirow{2}{*}{Method} & \multirow{2}{*}{Online}
      & \multicolumn{3}{c}{Sintel (50 frames)}
      & \multicolumn{3}{c}{TUM-dynamics (90 frames)} \\
    \cmidrule(lr){3-5} \cmidrule(lr){6-8}
      & & ATE $\downarrow$ & RPE trans $\downarrow$ & RPE rot $\downarrow$
      & ATE $\downarrow$ & RPE trans $\downarrow$ & RPE rot $\downarrow$ \\
    \midrule
    VGGT~\cite{wang2025vggt}      & \ding{55} & 0.206 & 0.129 & 1.893 & \textbf{0.006} & \underline{0.008} & 1.435 \\
    \midrule
    Spann3R~\cite{wang20253d}   & \ding{51} & 0.329 & 0.162 & 5.328 & 0.042 & 0.020 & 1.055 \\
    CUT3R~\cite{wang2025continuous}     & \ding{51} & \underline{0.189} & \textbf{0.058} & \textbf{0.535} & 0.029 & 0.008 & 0.271 \\
    Point3R~\cite{wu2025point3r}   & \ding{51} & 0.384 & 0.133 & 1.539 & 0.036 & 0.017 & 0.476 \\
    \rowcolor{gray!15} StreamVGGT$^\dagger$~\cite{zhuo2025streaming} & \ding{51} & 0.227 & 0.130 & 1.667 & 0.019 & 0.012 & 0.923 \\
    STream3R$\beta$~\cite{lan2025stream3r} & \ding{51} & \textbf{0.184} & \underline{0.063} & 0.639 & \underline{0.012} & 0.008 & \textbf{0.234} \\
    InfiniteVGGT~\cite{yuan2026infinitevggt} & \ding{51} & 0.227 & 0.130 & 1.667 & 0.020 & 0.012 & 0.926 \\
    TTT3R~\cite{chen2025ttt3r}     & \ding{51} & 0.190 & 0.073 & \underline{0.592} & 0.017 & \textbf{0.007} & \underline{0.254} \\
    \rowcolor{cyan!8} RetrieveVGGT (Ours) & \ding{51} & 0.227 & 0.130 & 1.667 & 0.018 & 0.012 & 0.923 \\
    \bottomrule
    \end{tabular}%
    }
\end{table*}

\begin{table*}[!t]
    \centering
    \caption{\textbf{Video depth estimation on short sequences.} We report scale-invariant relative depth (Per-sequence Scale) on Sintel~\cite{butler2012naturalistic} (50 frames), Bonn~\cite{palazzolo2019refusion} (110 frames), and KITTI~\cite{geiger2013vision} (110 frames). Best results are in \textbf{bold}, second best are \underline{underlined}. $^\dagger$Base model of RetrieveVGGT.}
    \label{tab:supp_depth}
    \resizebox{\textwidth}{!}{%
    \begin{tabular}{lc cc cc cc}
    \toprule
    \multirow{2}{*}{Method} & \multirow{2}{*}{Online}
      & \multicolumn{2}{c}{Sintel (50 frames)}
      & \multicolumn{2}{c}{Bonn (110 frames)}
      & \multicolumn{2}{c}{KITTI (110 frames)} \\
    \cmidrule(lr){3-4} \cmidrule(lr){5-6} \cmidrule(lr){7-8}
      & & Abs Rel $\downarrow$ & $\delta{<}1.25$ $\uparrow$
      & Abs Rel $\downarrow$ & $\delta{<}1.25$ $\uparrow$
      & Abs Rel $\downarrow$ & $\delta{<}1.25$ $\uparrow$ \\
    \midrule
    VGGT~\cite{wang2025vggt}      & \ding{55} & \underline{0.287} & \underline{66.1} & \textbf{0.055} & 97.1 & \textbf{0.070} & \textbf{96.5} \\
    \midrule
    Spann3R~\cite{wang20253d}   & \ding{51} & 0.622 & 42.6 & 0.144 & 81.3 & 0.198 & 73.7 \\
    CUT3R~\cite{wang2025continuous}     & \ding{51} & 0.421 & 47.9 & 0.078 & 93.7 & 0.118 & 88.1 \\
    Point3R~\cite{wu2025point3r}   & \ding{51} & 0.452 & 48.9 & 0.060 & 96.0 & 0.136 & 84.2 \\
    \rowcolor{gray!15} StreamVGGT$^\dagger$~\cite{zhuo2025streaming} & \ding{51} & 0.323 & 65.7 & 0.059 & 97.2 & 0.1753 & 72.1 \\
    STream3R$\beta$~\cite{lan2025stream3r} & \ding{51} & \textbf{0.265} & \textbf{70.5} & 0.070 & 95.2 & \underline{0.079} & \underline{94.8} \\
    InfiniteVGGT~\cite{yuan2026infinitevggt} & \ding{51} & 0.325 & 65.6 & 0.060 & \underline{97.3} & 0.1753 & 72.4 \\
    TTT3R~\cite{chen2025ttt3r}     & \ding{51} & 0.404 & 50.0 & 0.068 & 95.4 & 0.113 & 90.4 \\
    \rowcolor{cyan!8} RetrieveVGGT (Ours) & \ding{51} & 0.325 & 65.8 & \underline{0.057} & \textbf{97.4} & 0.1754 & 71.8 \\
    \bottomrule
    \end{tabular}%
    }
\end{table*}

\section{Qualitative 3D Reconstruction Results}
\label{sec:qual_3d}

\begin{figure*}[p]
\centering
\setlength{\tabcolsep}{2pt}
\renewcommand{\arraystretch}{0.5}

\begin{tabular}{cccc}
{\small Reference} & {\small TTT3R~\cite{chen2025ttt3r}} & {\small \textbf{RetrieveVGGT (Ours)}} & {\small Ground Truth} \\[4pt]

\includegraphics[height=0.16\linewidth]{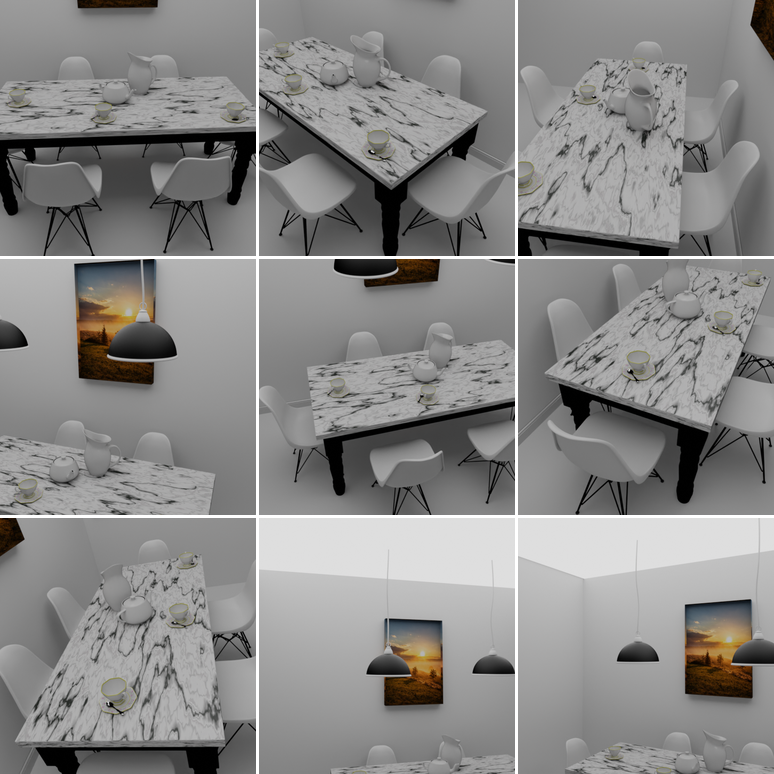} &
\includegraphics[height=0.16\linewidth]{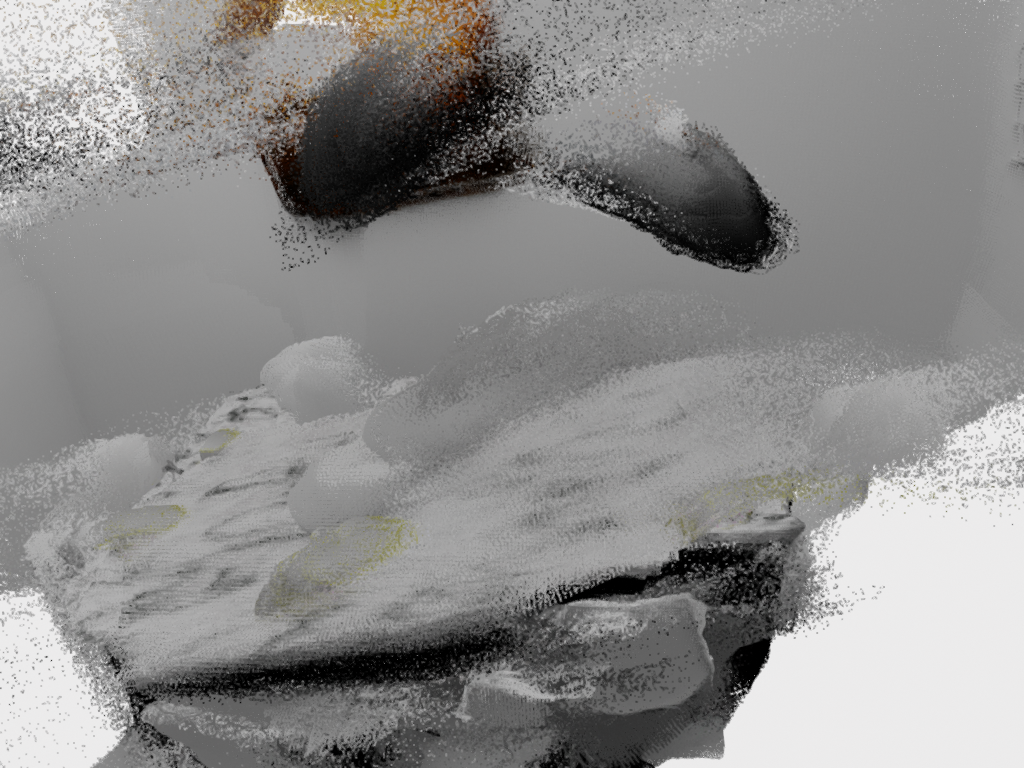} &
\includegraphics[height=0.16\linewidth]{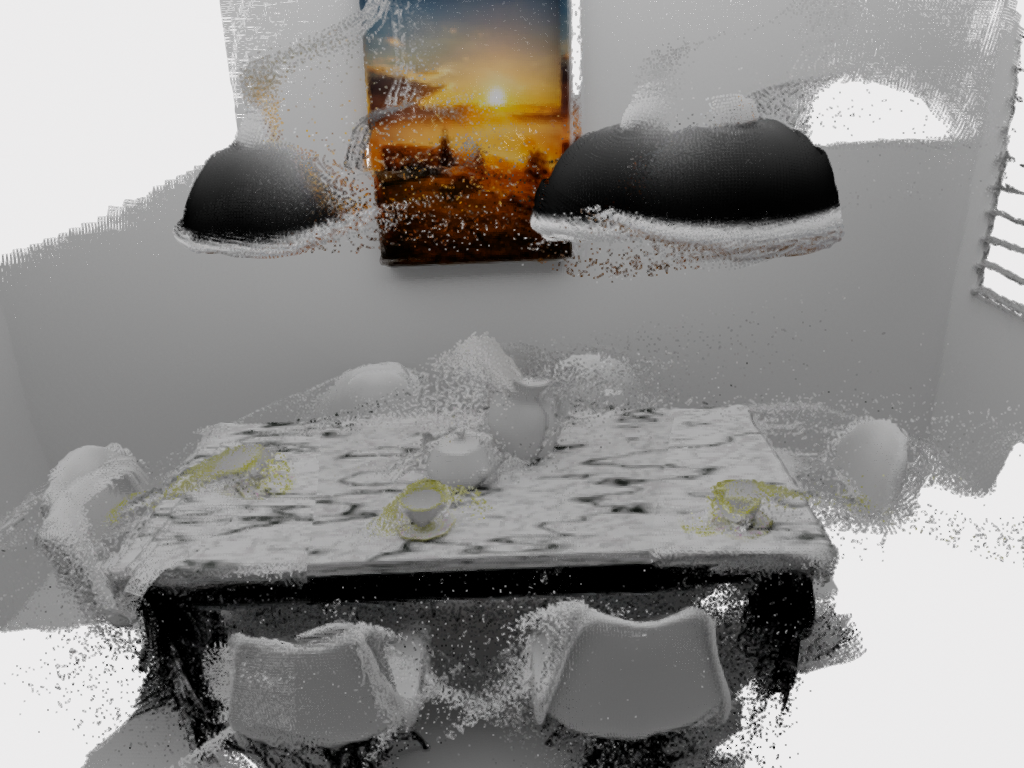} &
\includegraphics[height=0.16\linewidth]{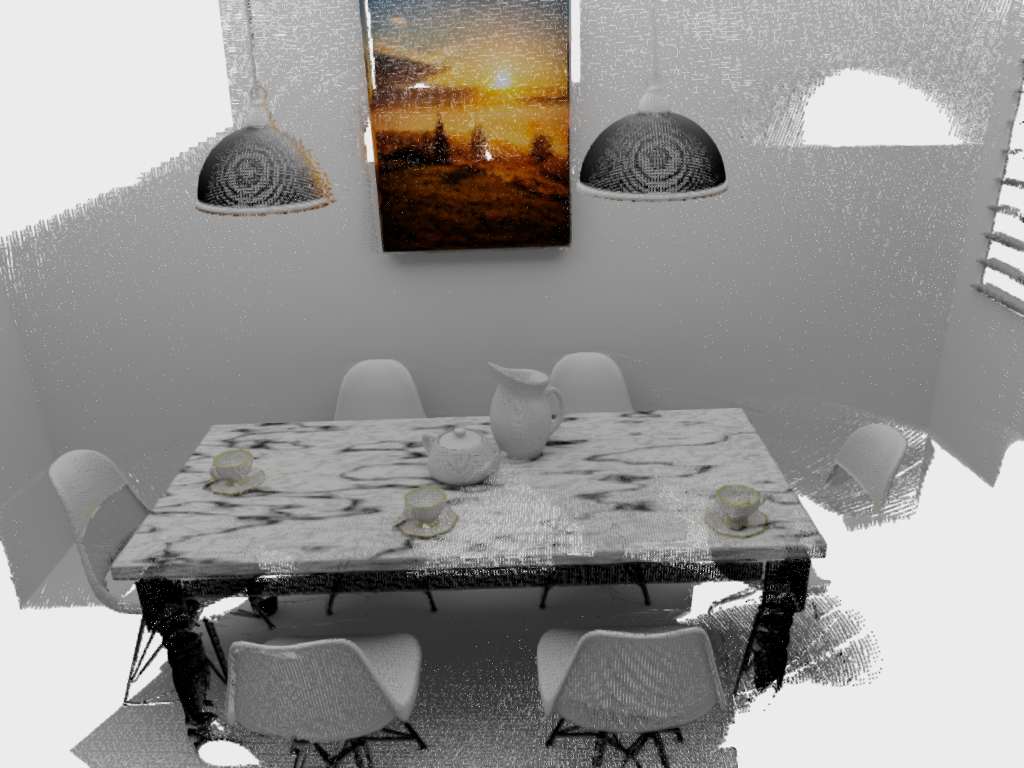} \\[-2pt]
\multicolumn{4}{c}{\scriptsize NRGBD: \texttt{breakfast\_room}} \\[4pt]

\includegraphics[height=0.16\linewidth]{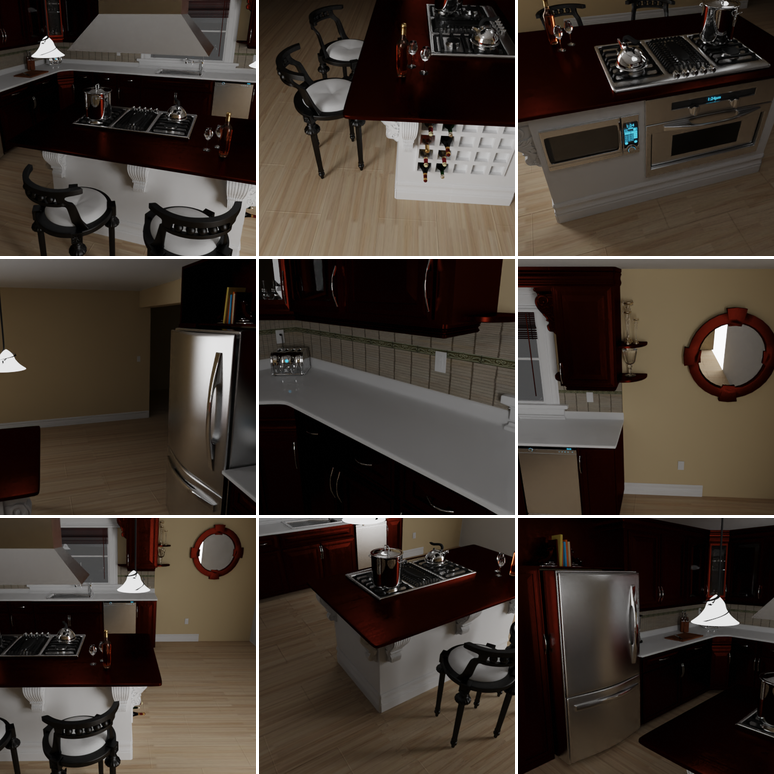} &
\includegraphics[height=0.16\linewidth]{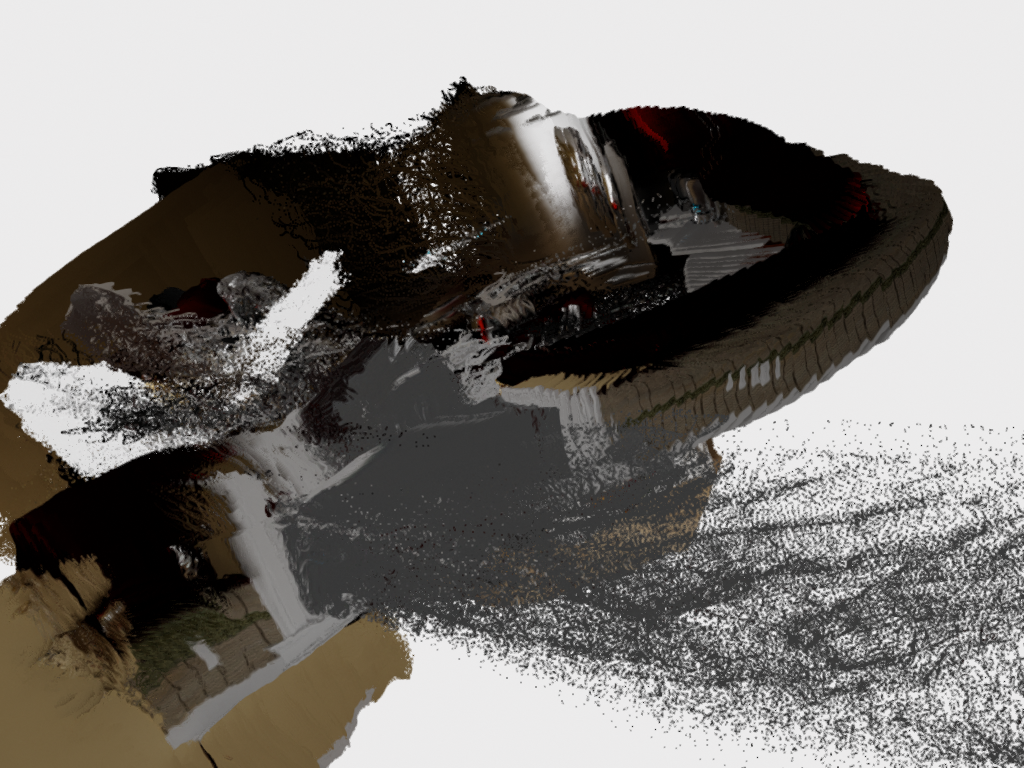} &
\includegraphics[height=0.16\linewidth]{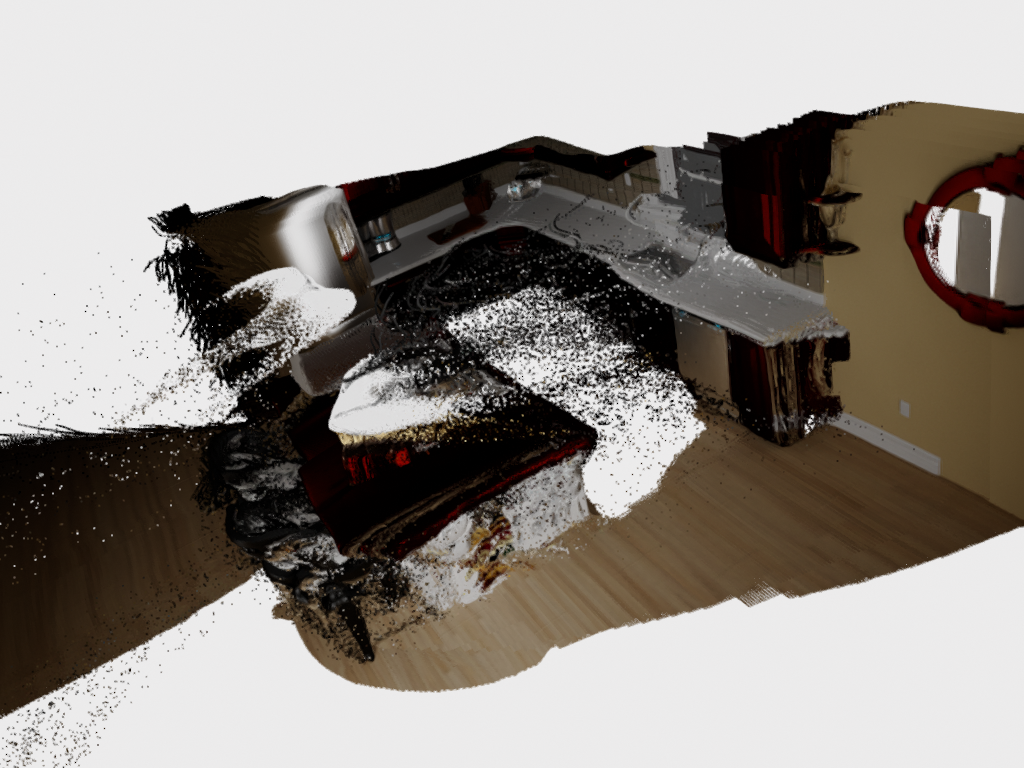} &
\includegraphics[height=0.16\linewidth]{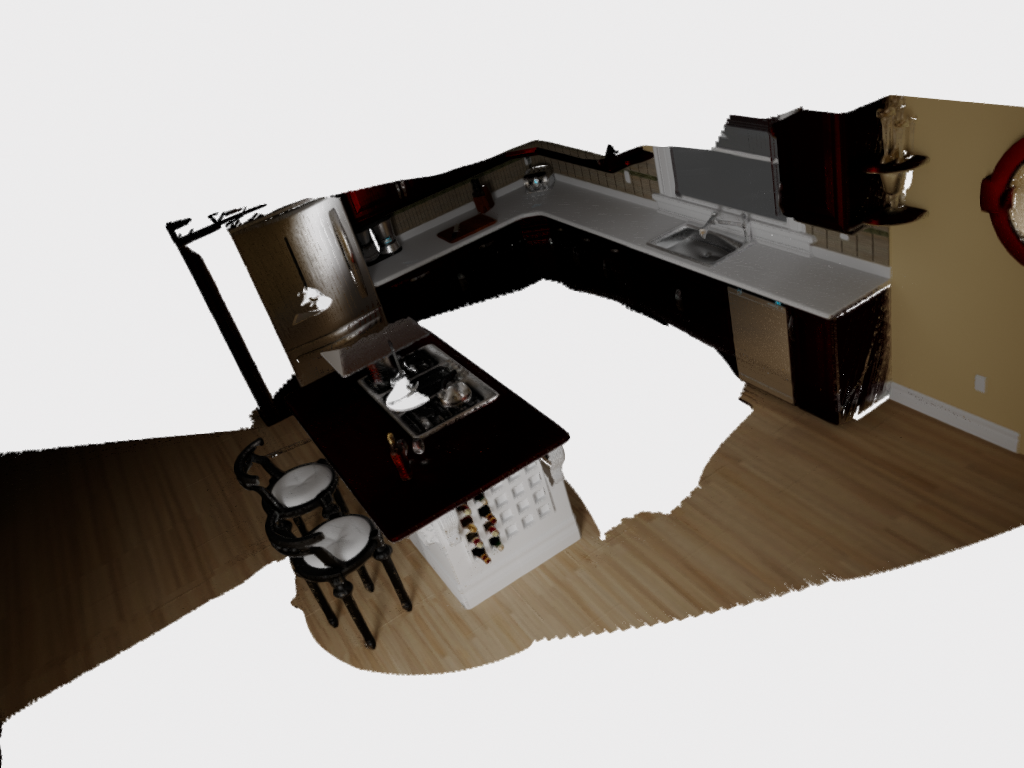} \\[-2pt]
\multicolumn{4}{c}{\scriptsize NRGBD: \texttt{complete\_kitchen}} \\[4pt]

\includegraphics[height=0.16\linewidth]{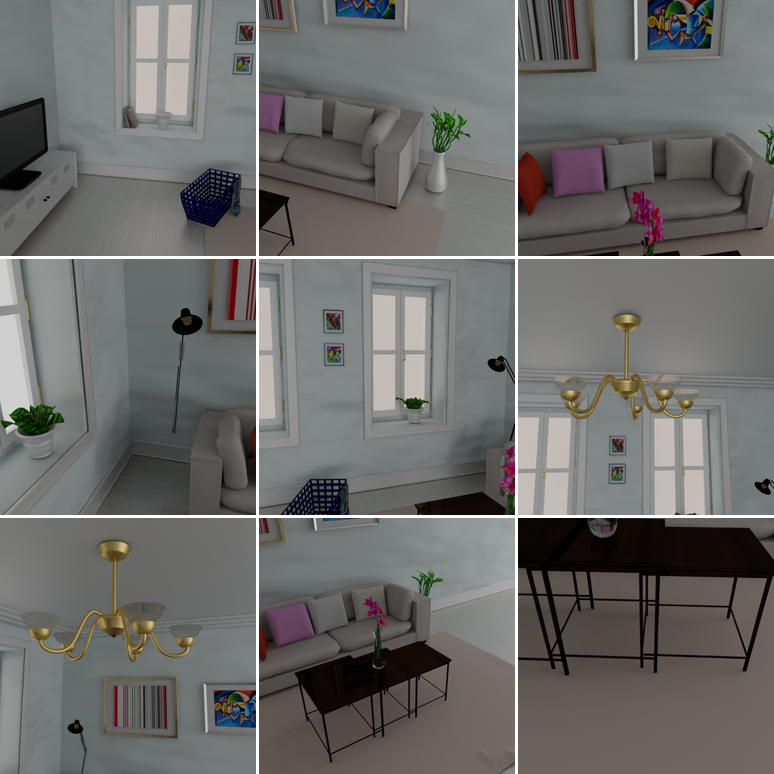} &
\includegraphics[height=0.16\linewidth]{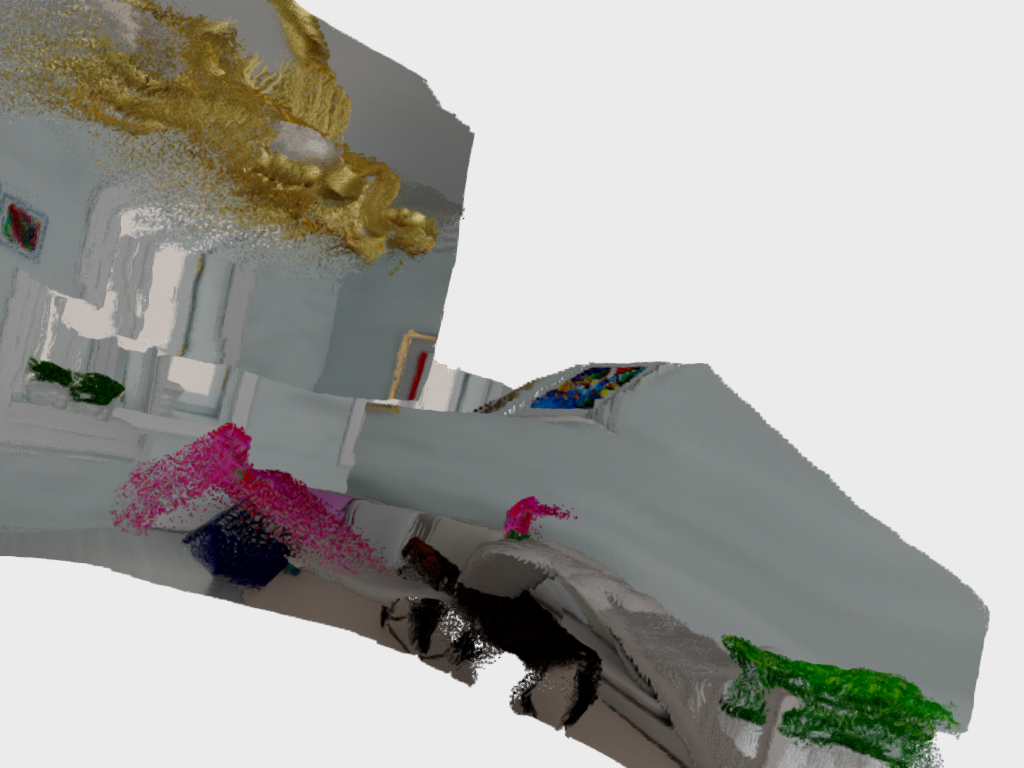} &
\includegraphics[height=0.16\linewidth]{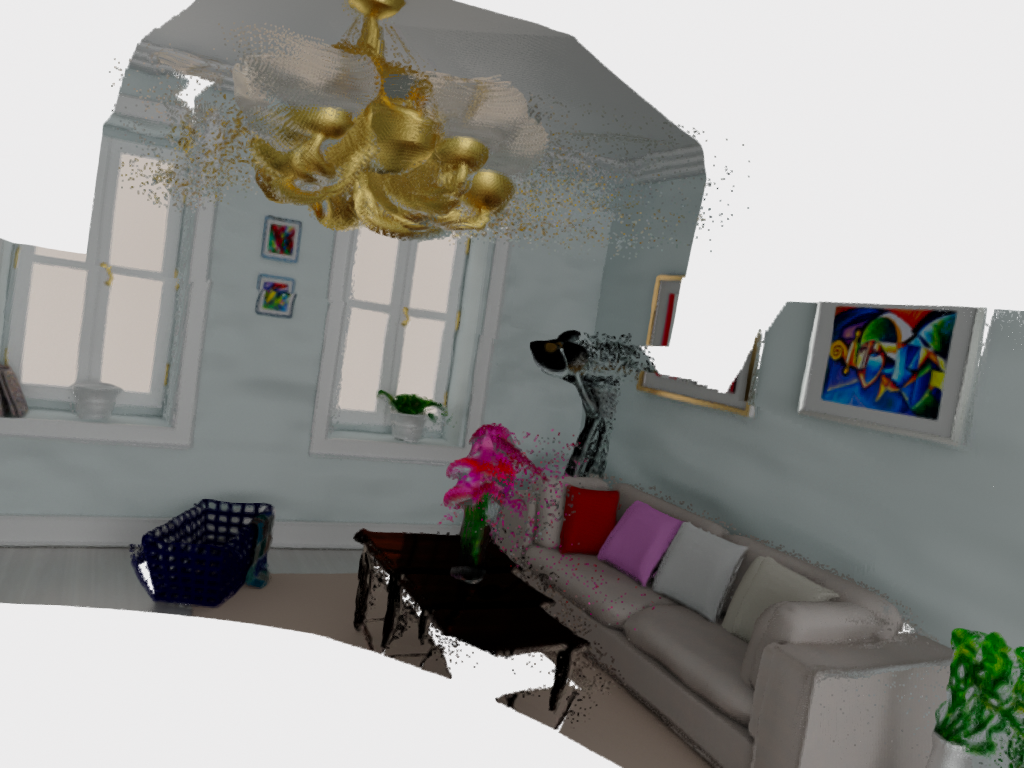} &
\includegraphics[height=0.16\linewidth]{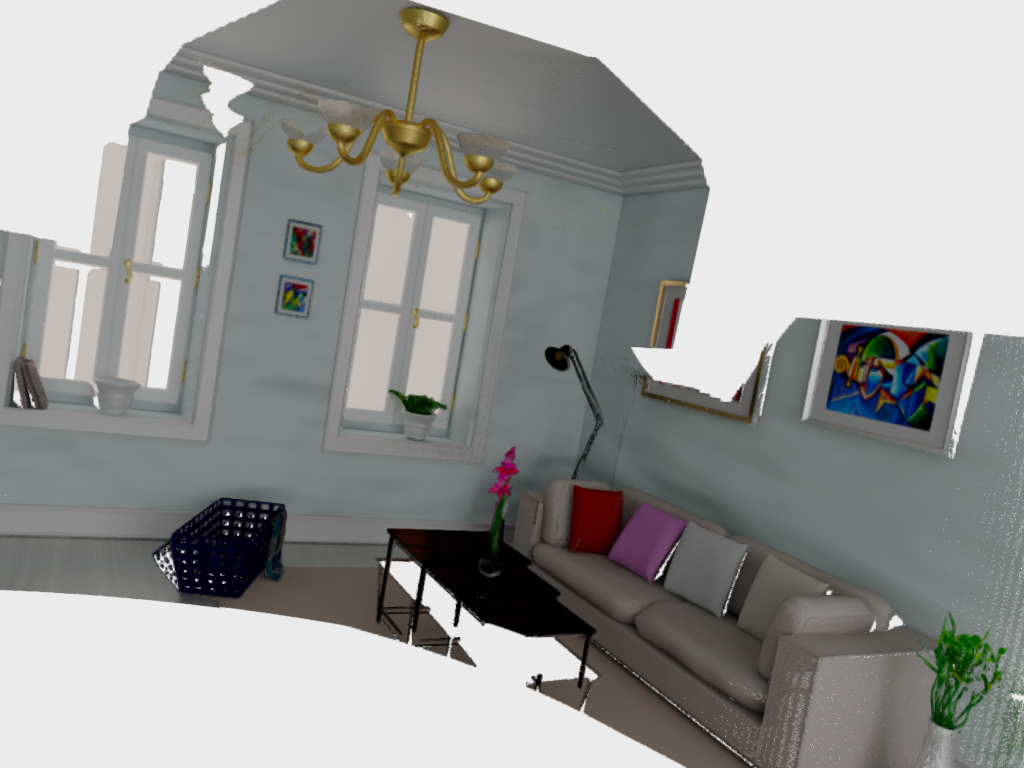} \\[-2pt]
\multicolumn{4}{c}{\scriptsize NRGBD: \texttt{green\_room}} \\[4pt]

\includegraphics[height=0.16\linewidth]{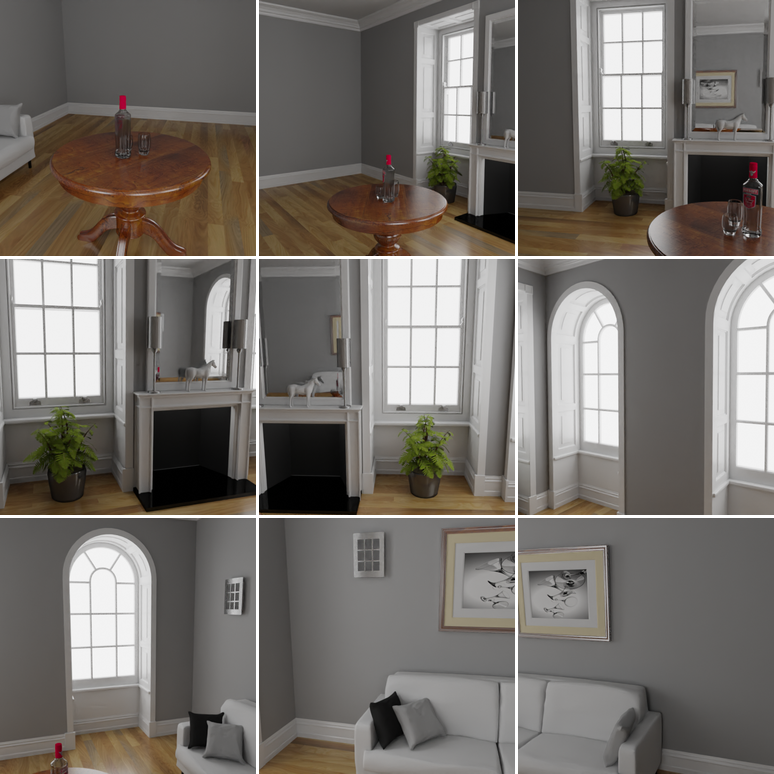} &
\includegraphics[height=0.16\linewidth]{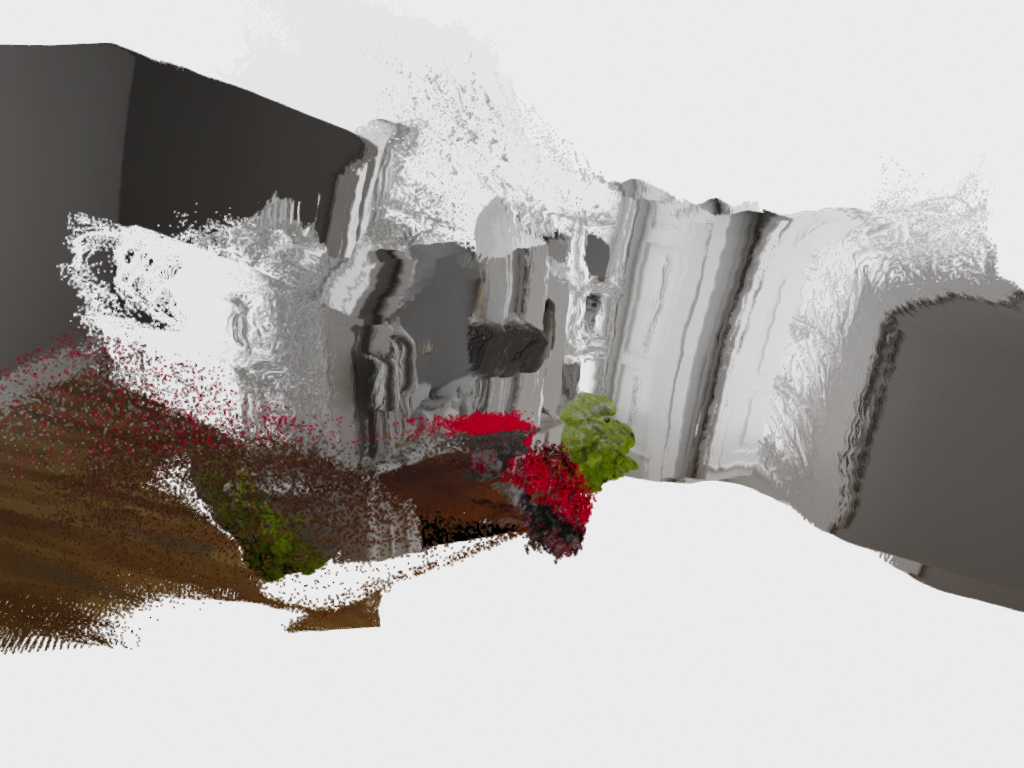} &
\includegraphics[height=0.16\linewidth]{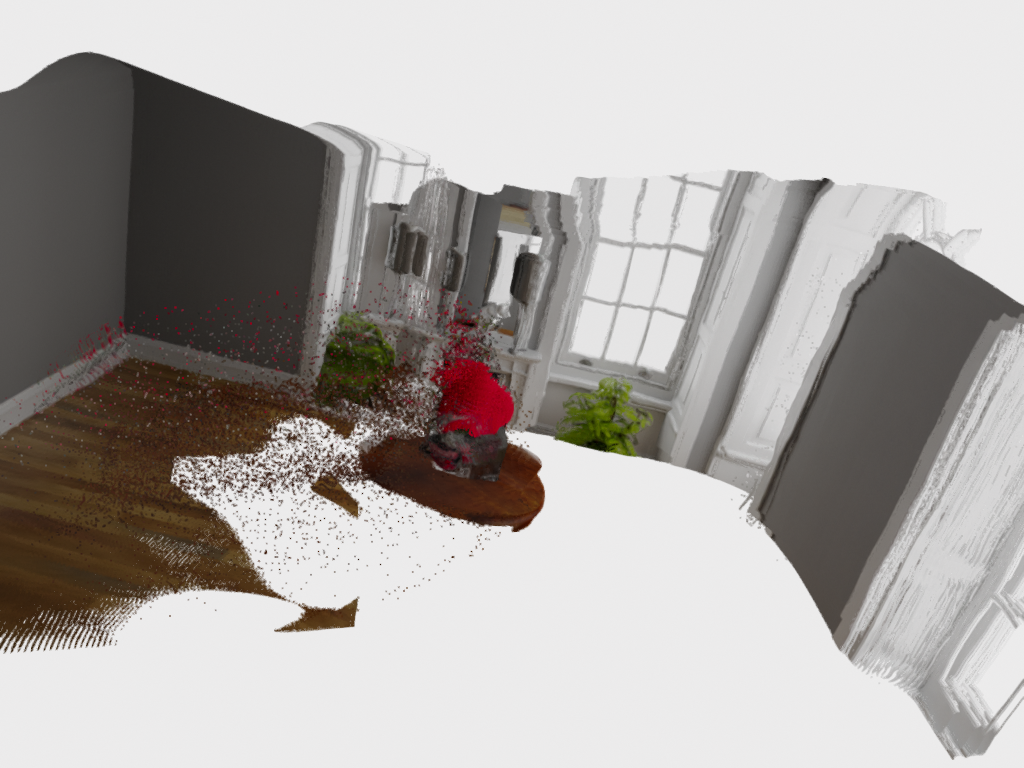} &
\includegraphics[height=0.16\linewidth]{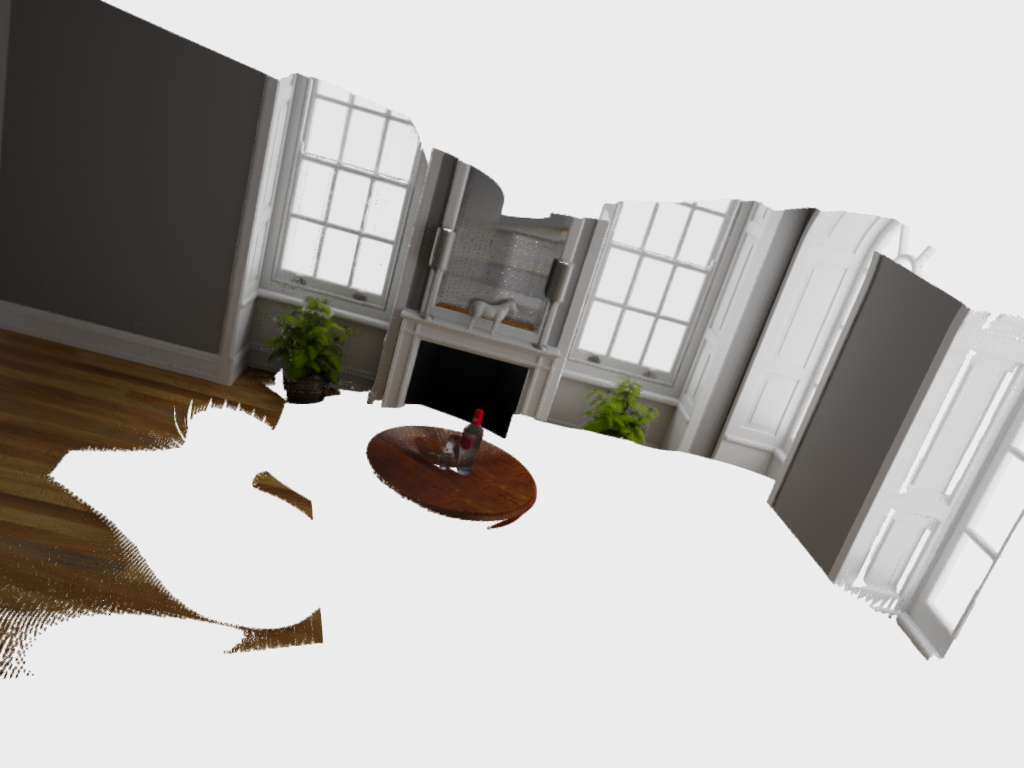} \\[-2pt]
\multicolumn{4}{c}{\scriptsize NRGBD: \texttt{grey\_white\_room}} \\[4pt]

\includegraphics[height=0.16\linewidth]{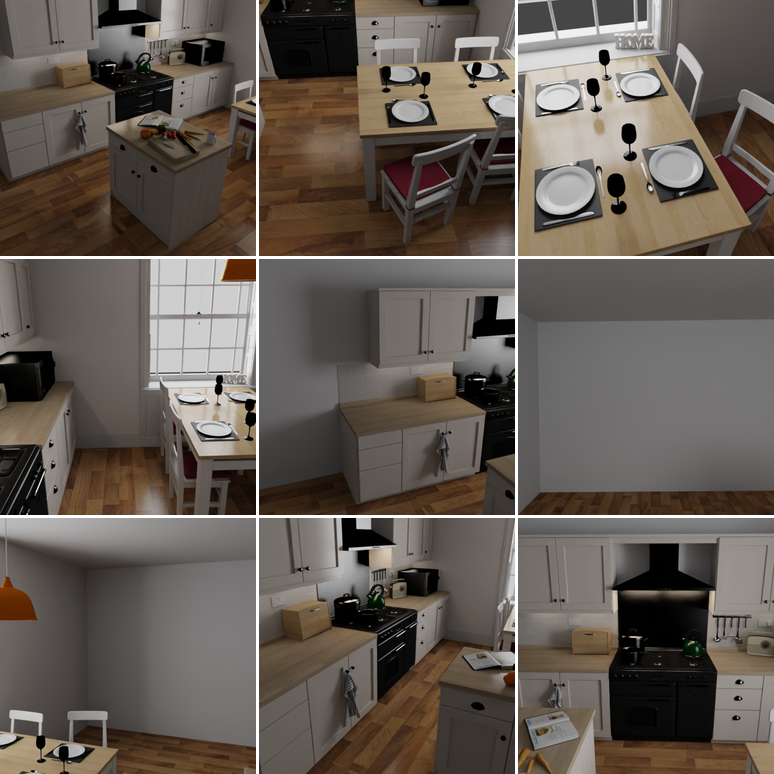} &
\includegraphics[height=0.16\linewidth]{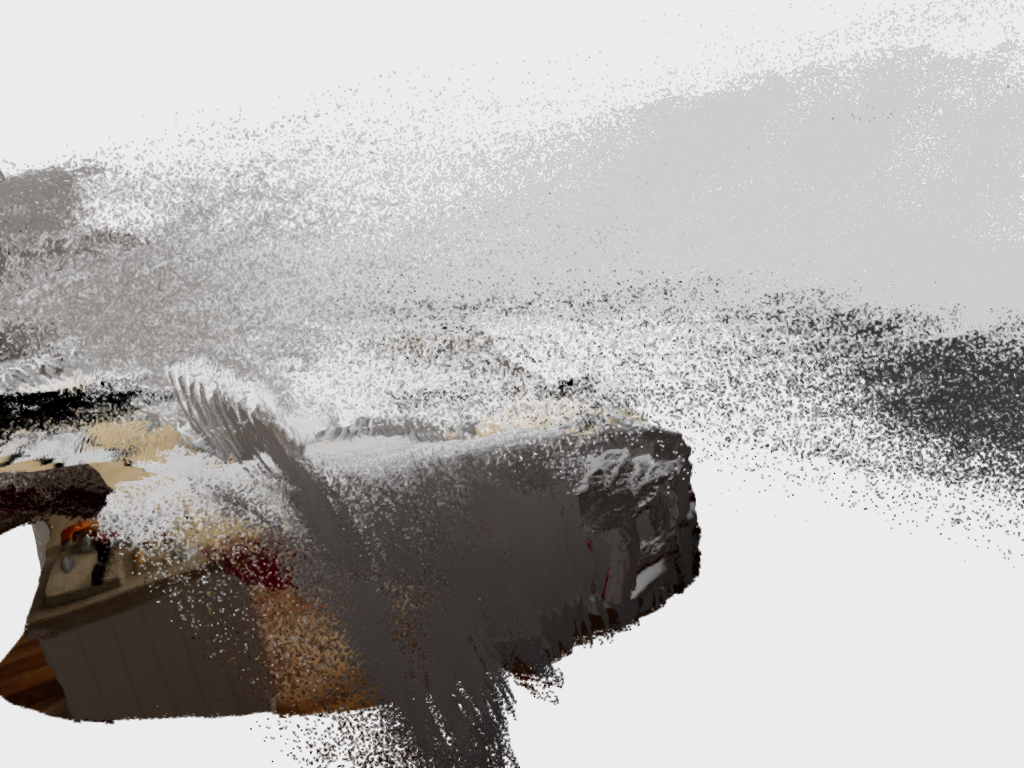} &
\includegraphics[height=0.16\linewidth]{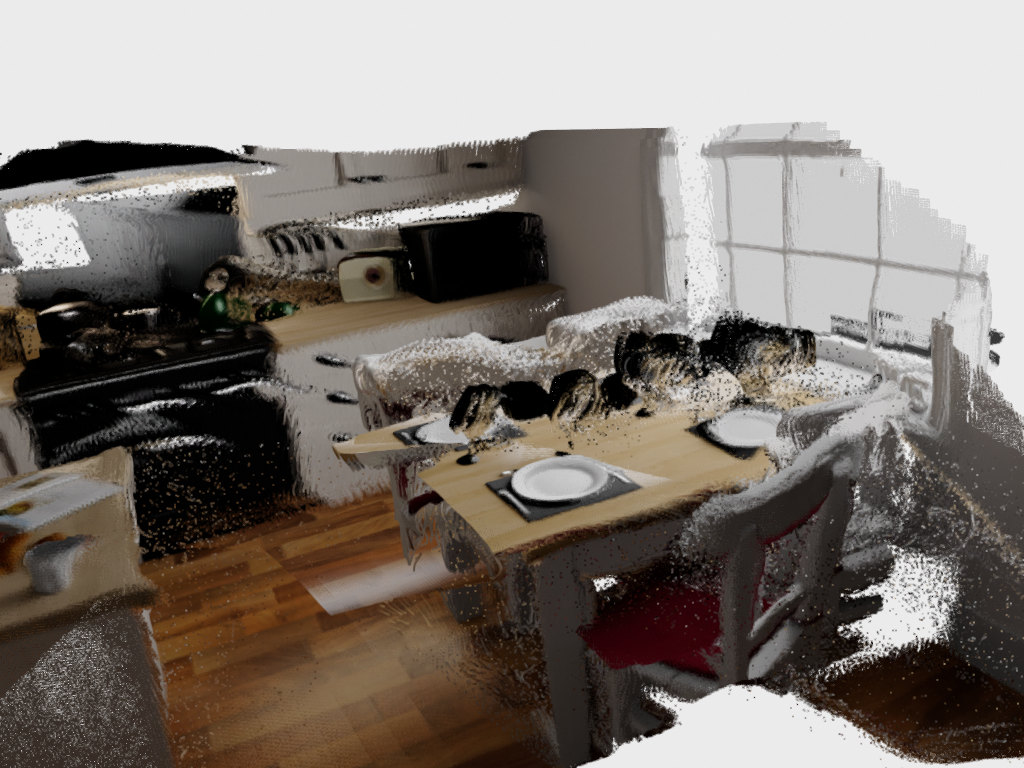} &
\includegraphics[height=0.16\linewidth]{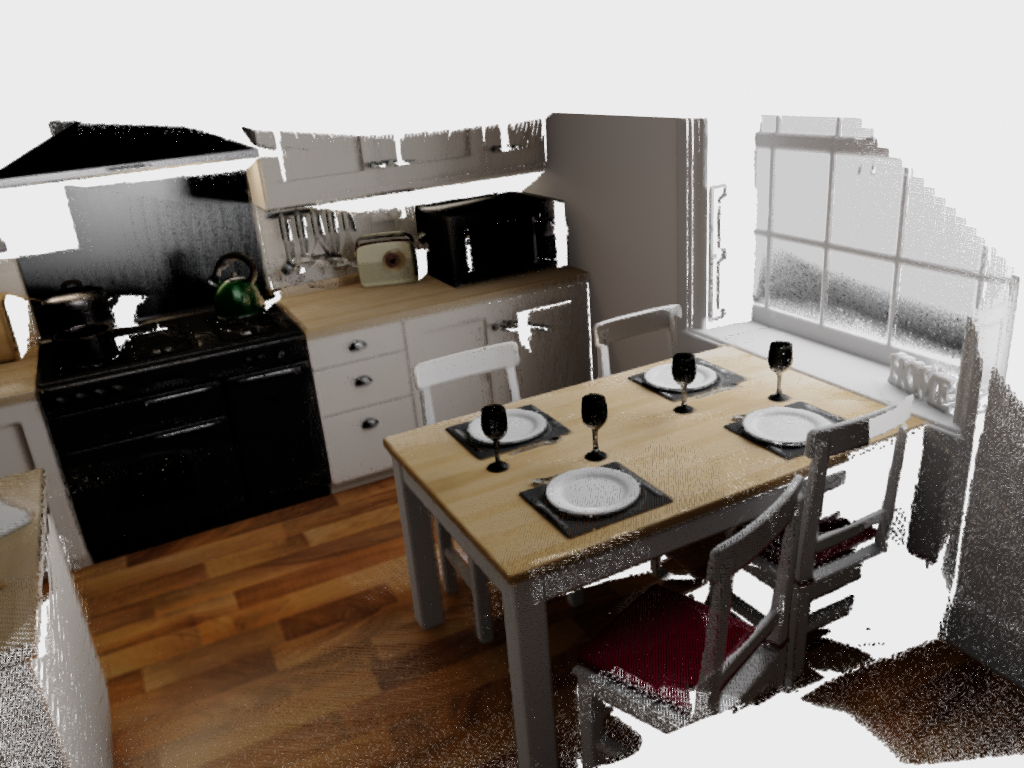} \\[-2pt]
\multicolumn{4}{c}{\scriptsize NRGBD: \texttt{kitchen}} \\[4pt]

\includegraphics[height=0.16\linewidth]{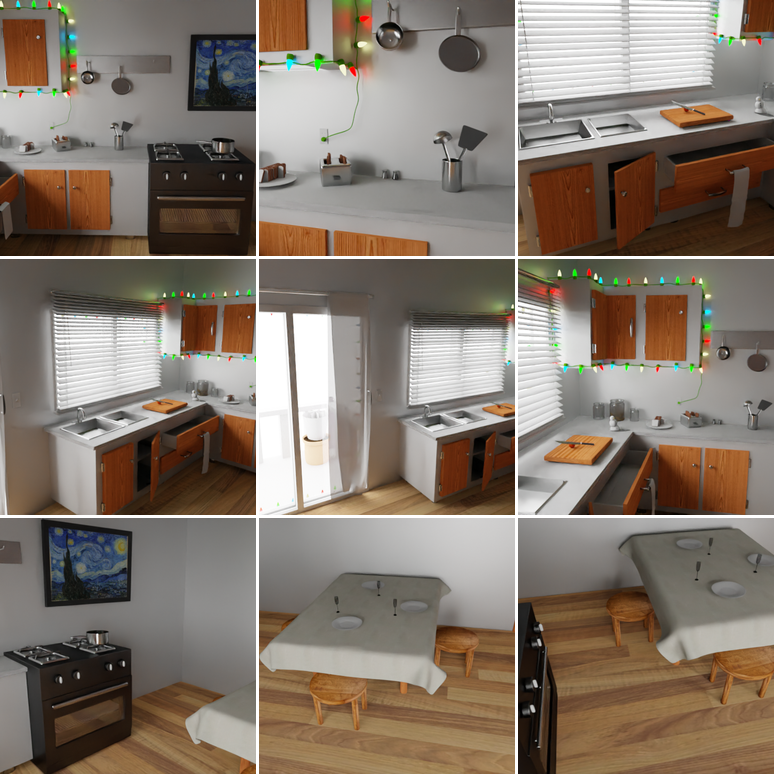} &
\includegraphics[height=0.16\linewidth]{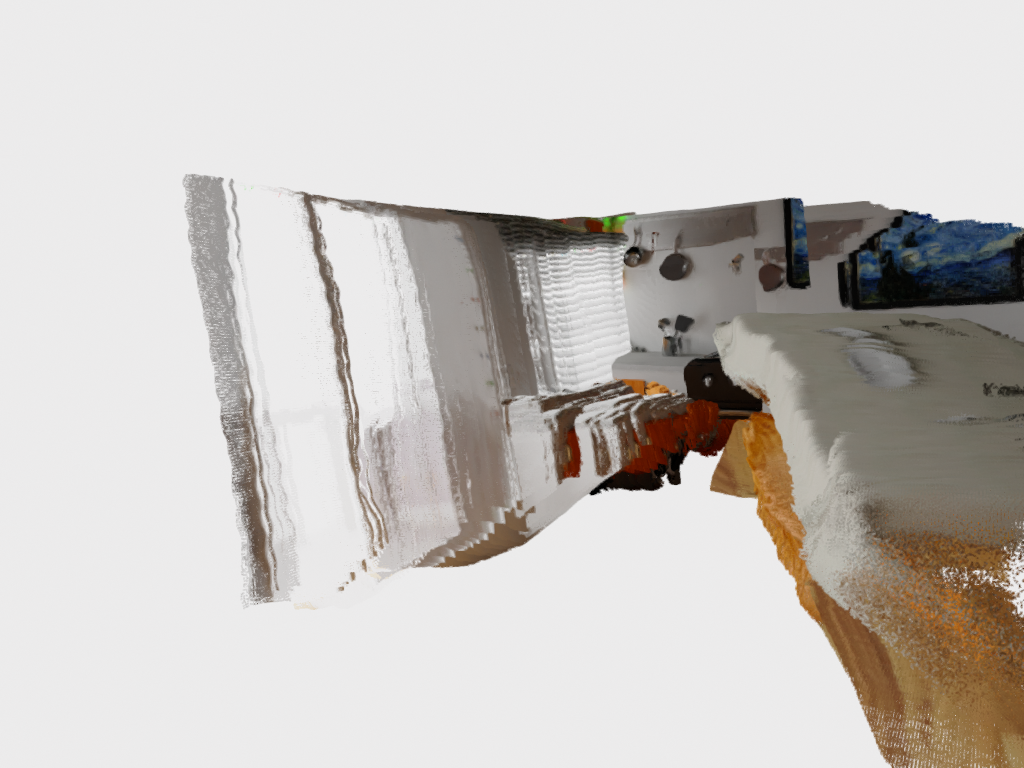} &
\includegraphics[height=0.16\linewidth]{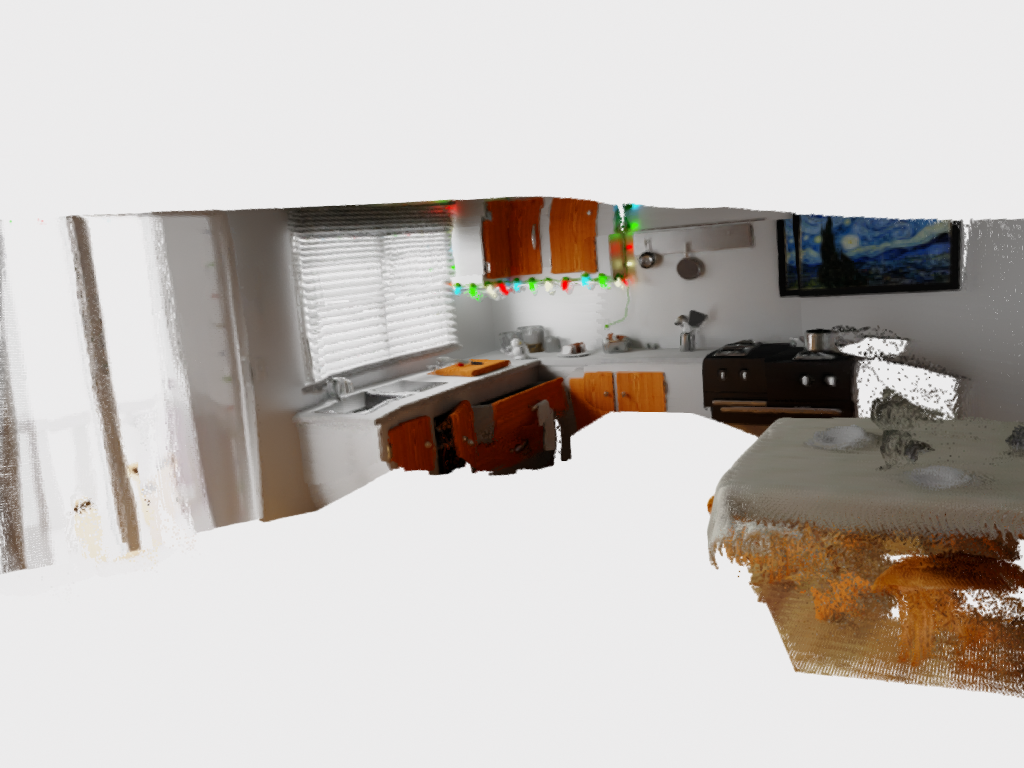} &
\includegraphics[height=0.16\linewidth]{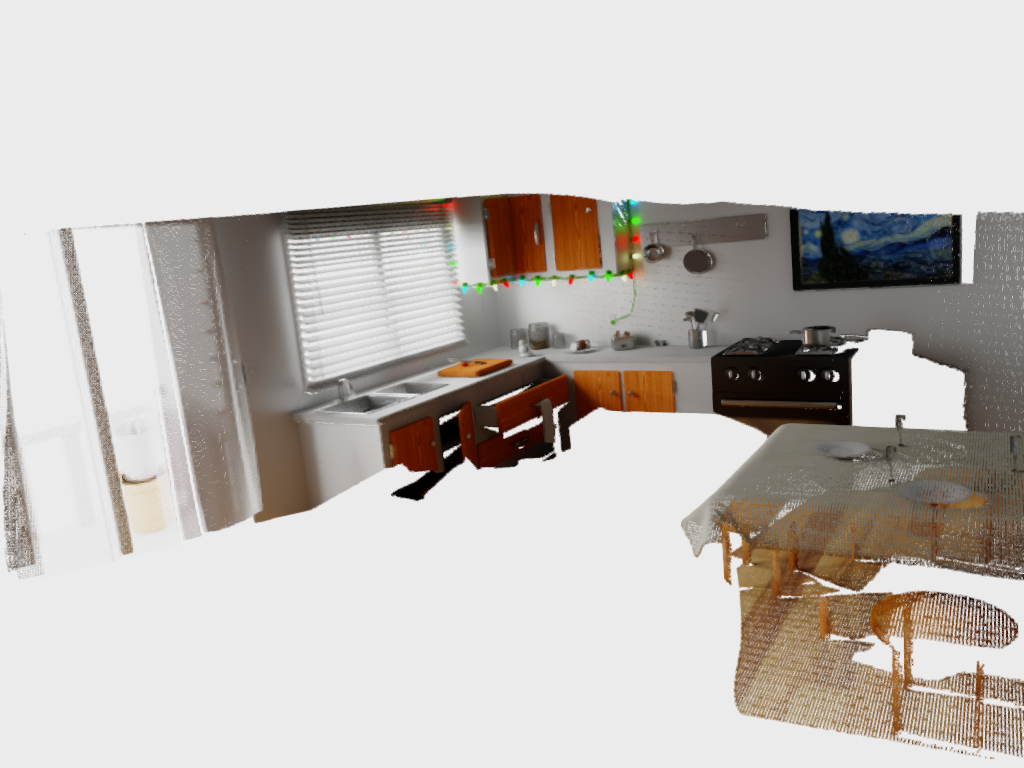} \\[-2pt]
\multicolumn{4}{c}{\scriptsize NRGBD: \texttt{morning\_apartment}} \\[4pt]

\includegraphics[height=0.16\linewidth]{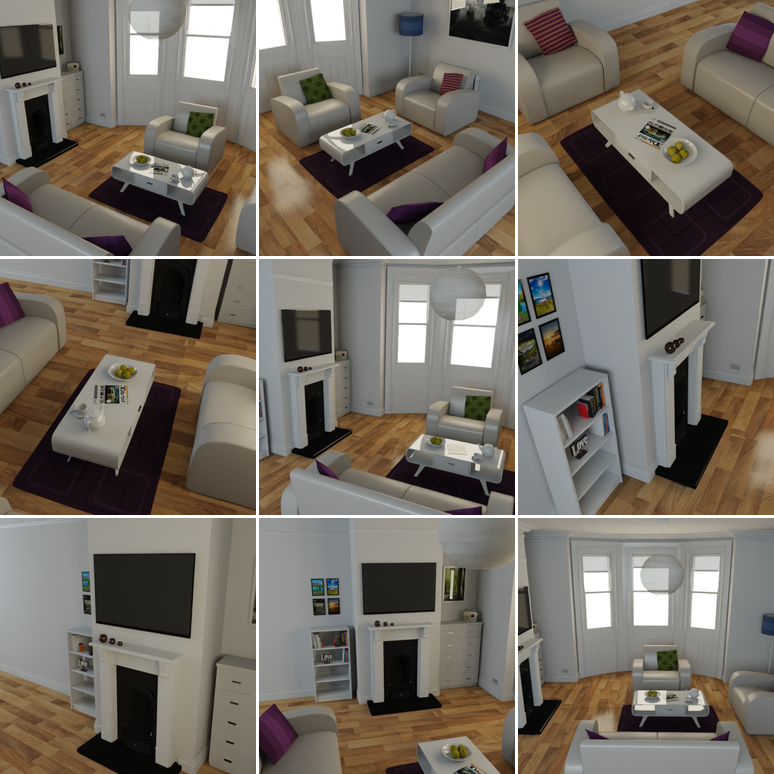} &
\includegraphics[height=0.16\linewidth]{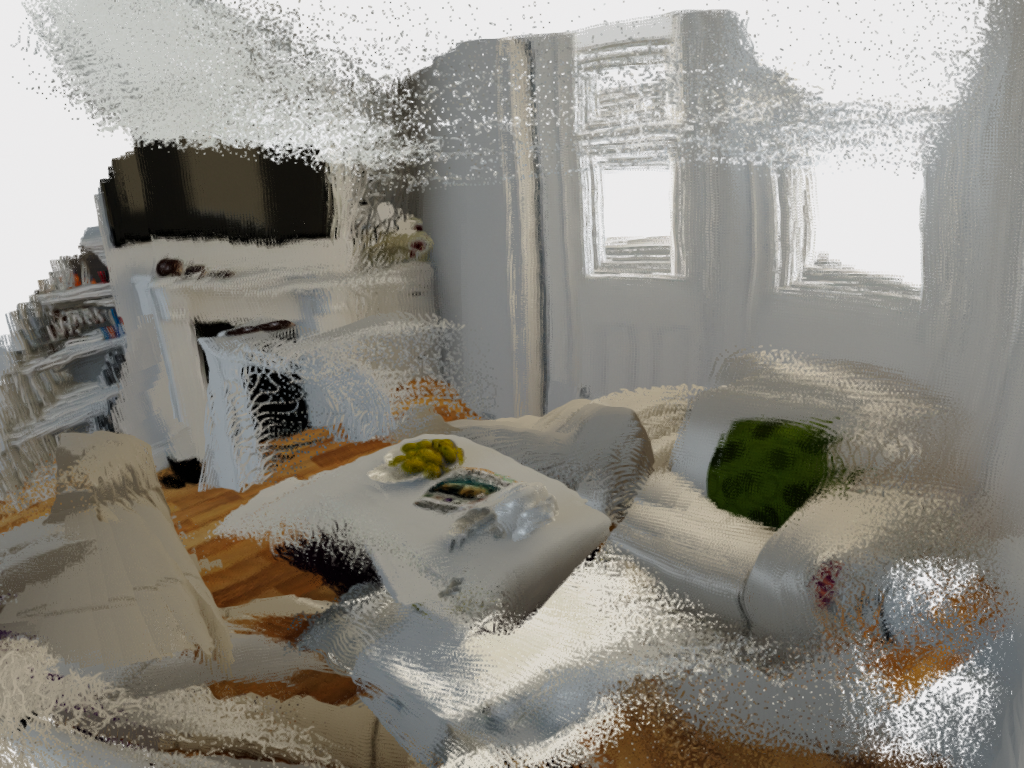} &
\includegraphics[height=0.16\linewidth]{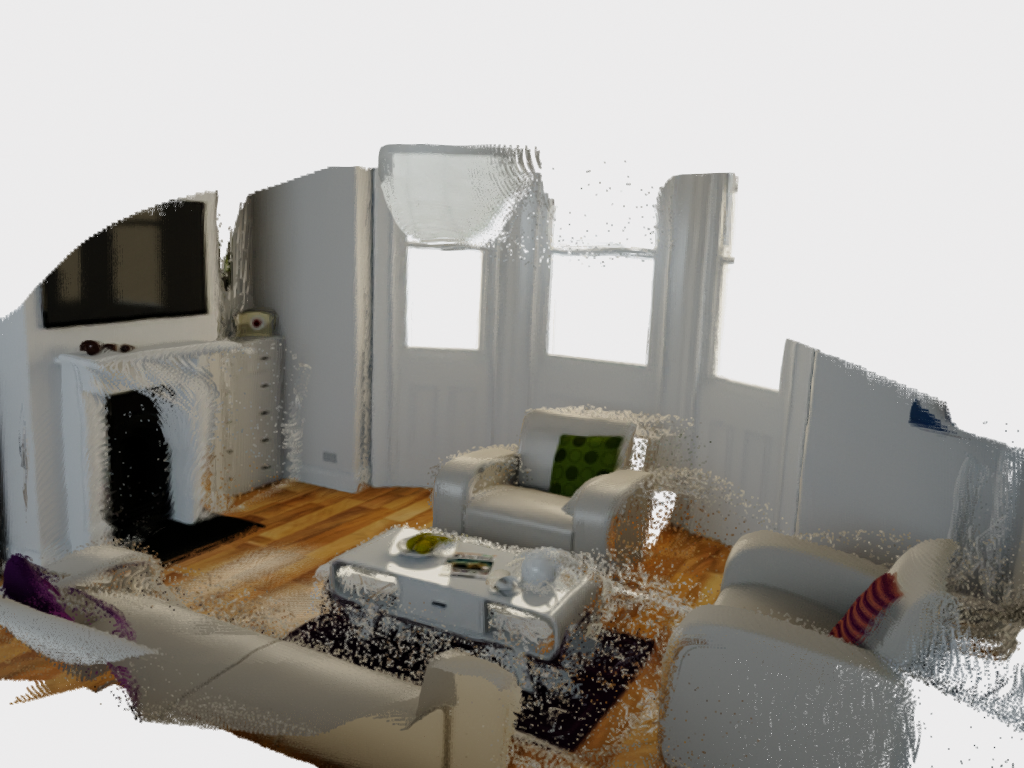} &
\includegraphics[height=0.16\linewidth]{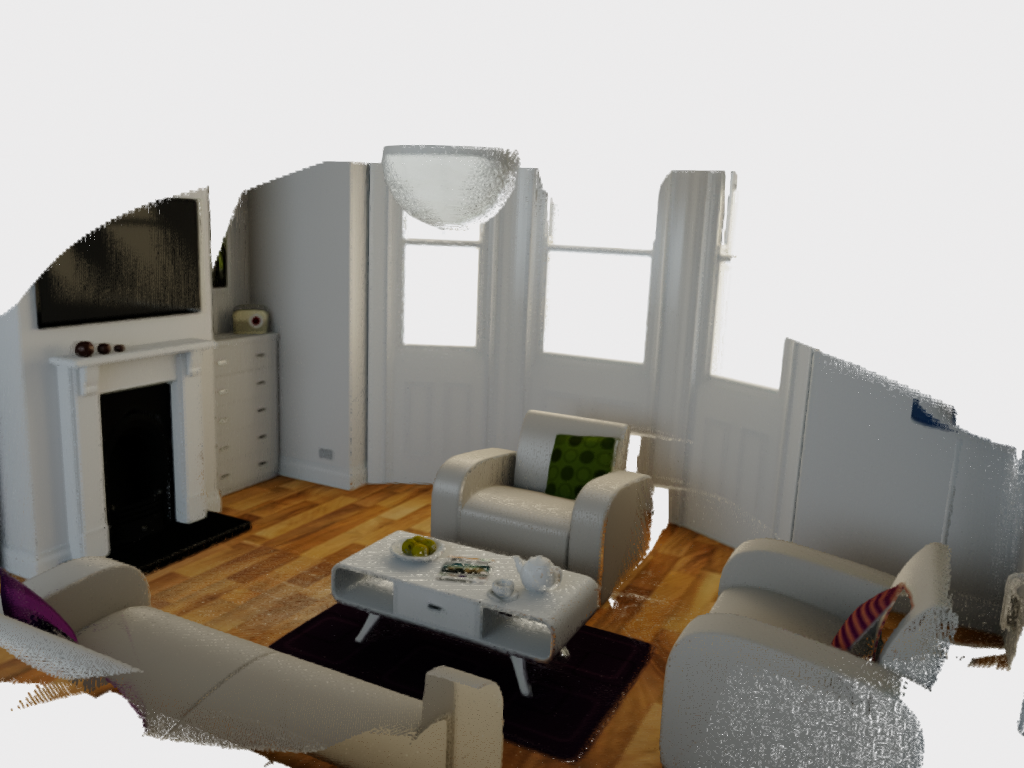} \\[-2pt]
\multicolumn{4}{c}{\scriptsize NRGBD: \texttt{whiteroom}} \\

\end{tabular}

\caption{\textbf{Qualitative 3D reconstruction on NRGBD.} Point cloud reconstructions on NRGBD scenes, all visualized from an identical viewpoint.}
\label{fig:qual3d_merged}
\end{figure*}

Fig.~\ref{fig:qual3d_merged} shows qualitative comparisons on NRGBD. For each scene, we show reference images, predictions from TTT3R and RetrieveVGGT, and ground truth from an identical viewpoint. RetrieveVGGT maintains structural fidelity without the distortion or catastrophic forgetting seen in recurrent methods, particularly in large-scale scenes where long-range context is critical.

\begin{figure}[tb]
    \centering
    \includegraphics[width=1.0\linewidth]{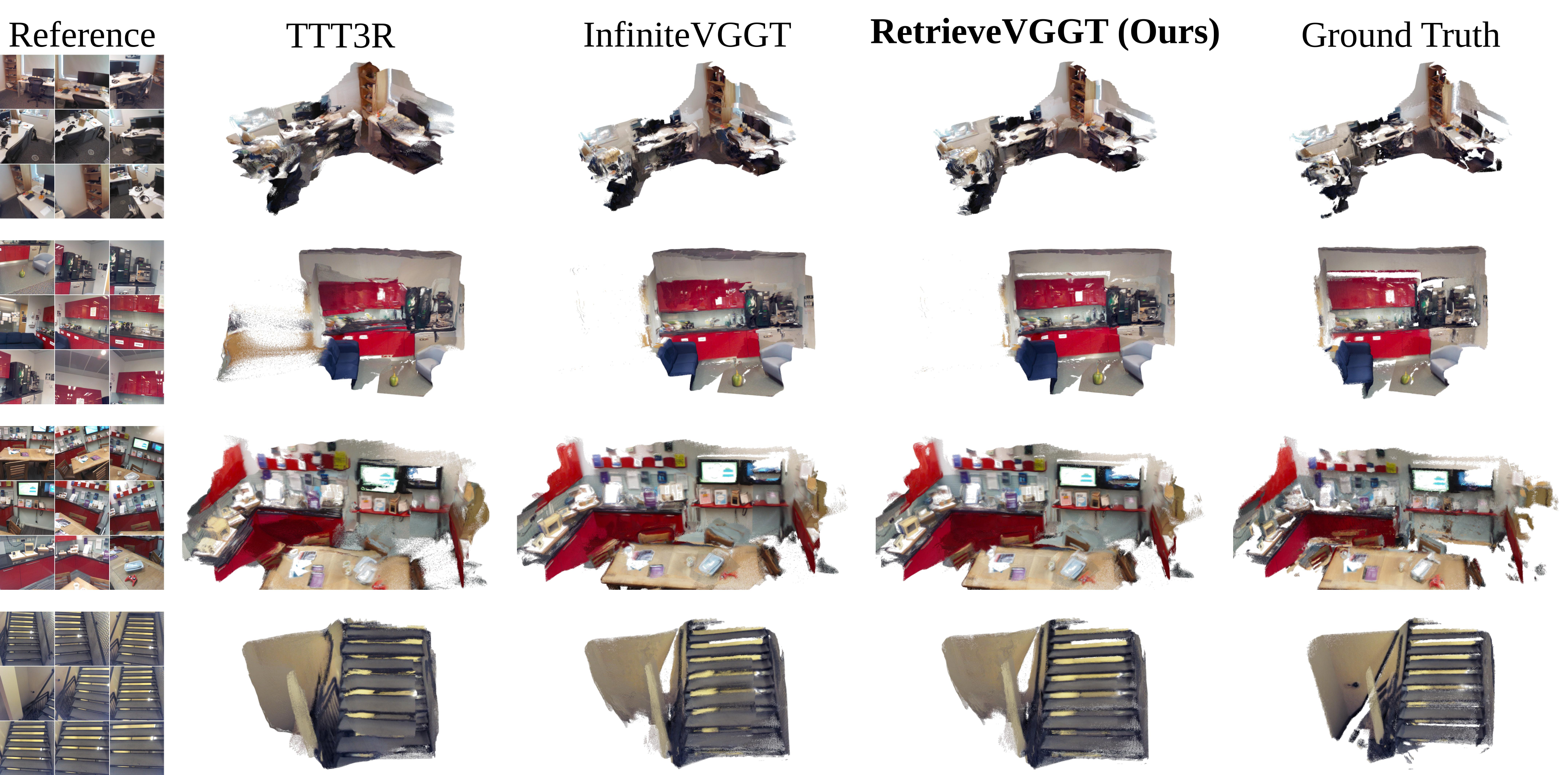}
    \caption{\textbf{Qualitative Comparison of 3D reconstruction on 7Scenes.} Compared with TTT3R and InfiniteVGGT, RetrieveVGGT (Ours) produces more complete and geometrically consistent reconstructions with fewer outliers.}
    \label{fig:qual3d_main}
\end{figure}

Fig.~\ref{fig:qual3d_main} compares 3D reconstructions on 7-Scenes. RetrieveVGGT yields more complete geometry with fewer outliers than both TTT3R and InfiniteVGGT.

\section{Limitations}
\label{sec:limitations}

As a training-free framework, RetrieveVGGT inherits the capability boundaries of its base model (e.g., dynamic objects, textureless regions). The fixed retrieval budget $N$ does not adapt to scene complexity. Evaluation is limited to indoor scenes; highly dynamic environments remain to be validated.

\section{Broader Impacts}
\label{sec:broader}

RetrieveVGGT lowers the barrier for long-sequence 3D reconstruction on commodity GPUs, benefiting applications in robotics, AR/VR, and cultural heritage. As with any 3D reconstruction technology, potential misuse for unauthorized spatial mapping warrants careful attention to privacy guidelines.

\end{document}